\pdfoutput=1

\documentclass[11pt]{article}

\usepackage[final]{acl}

\usepackage{times}
\usepackage{latexsym}

\usepackage[T1]{fontenc}

\usepackage[utf8]{inputenc}

\usepackage{microtype}

\usepackage{inconsolata}

\usepackage{graphicx}

\usepackage{svg}
\usepackage{courier}
\usepackage{algorithm}
\usepackage{algorithmic}
\usepackage{soul}
\definecolor{prompt_color}{RGB}{242,242,242}
\usepackage{tcolorbox}
\usepackage{listings}
\tcbuselibrary{breakable}
\usepackage{multirow}
\usepackage[normalem]{ulem}
\useunder{\uline}{\ul}{}
\usepackage{array}
\usepackage{booktabs}
\usepackage{amssymb}
\usepackage{xspace}

\usepackage{enumitem}
\setenumerate[1]{itemsep=0pt,parsep=0pt,topsep=0pt}
\setitemize[1]{itemsep=5pt,parsep=0pt,topsep=5pt,leftmargin=10pt}

\usepackage{url} 

\usepackage{subfigure}

\newcommand{\textimage}[1]{\raisebox{-.1\height}{\includegraphics[height=1em]{#1}}}

\newcommand{\model}{\textbf{BAR}\xspace}

\definecolor{Step}{RGB}{0,112,192} 
\definecolor{Goal}{RGB}{32,153,85}

\soulregister\model7 
\soulregister\cite7
\soulregister\ref7 
\soulregister\citep7 
\soulregister\citet7 
\soulregister\pageref7 
\soulregister\name7 
\soulregister\xspace7 
\soulregister\underline7 
\soulregister\ding7 
\soulregister\textcircled7 
\soulregister\normalsize7
\soulregister\small7
\definecolor{achieve}{RGB}{0,153,0}

\makeatletter

\newcommand{\Rmnum}[1]{\expandafter\@slowromancap\romannumeral #1@}
\makeatother

\title{\model: A Backward Reasoning based Agent for Complex Minecraft Tasks}

\author{
{Weihong Du\textsuperscript{1}\textsuperscript{2}\quad Wenrui Liao\textsuperscript{1}\textsuperscript{2}\quad Binyu Yan\textsuperscript{1}\thanks{\quad Corresponding author} }
\\ 
{\textbf{Hongru Liang}\textsuperscript{\textbf{1}}\textsuperscript{\textbf{2}}\quad 
\textbf{Anthony G. Cohn}\textsuperscript{\textbf{3}}\textsuperscript{\textbf{4}}\quad \textbf{Wenqiang Lei}\textsuperscript{\textbf{1}}\textsuperscript{\textbf{2}}}
\\
{\textsuperscript{\textbf{1}}College of Computer Science, Sichuan University, China} \\
{\textsuperscript{\textbf{2}}Engineering Research Center of Machine Learning and Industry Intelligence,} \\
{Ministry of Education, China} \\
{\textsuperscript{\textbf{3}}School of Computer Science, University of Leeds, UK} \\
{\textsuperscript{\textbf{4}}The Alan Turing Institute, UK} \\
\texttt{\{duweihong, liaowenrui\}@stu.scu.edu.cn}\\
\texttt{\{yby, lianghongru, wenqianglei\}@scu.edu.cn}\\
\texttt{a.g.cohn@leeds.ac.uk} \\
}

\begin{document}
\maketitle

\begin{abstract}



Large language model~(LLM) based agents have shown great potential in following human instructions and automatically completing various tasks;
to do this, the agent needs to decompose it into easily executed steps by planning.
Existing LLM-based approaches to planning mostly proceed by inferring what steps should be inserted into the plan next by starting from the agent's initial state. 
However, this forward reasoning paradigm does not work well for complex tasks.
We study this issue in Minecraft, a virtual environment that simulates complex tasks based on real-world scenarios. The failure of forward reasoning is often caused by the large perception gap between the agent's initial state and task goal. 
To alleviate this, we leverage backward reasoning and make the planning start from the terminal (or goal) state, by first considering which actions could directly achieve the task goal in one step, before proceeding to consider how the preconditions of those actions can in turn be achieved. Our \underline{BA}ckward \underline{R}easoning based agent~(\model) is equipped with a recursive goal decomposition module, a state consistency maintaining module and a stage memory module. 
Experimental results demonstrate the superiority of \model over existing methods and the effectiveness of proposed modules. 
The code and dataset are available in \href{https://github.com/SCUNLP/BAR}{https://github.com/SCUNLP/BAR}. 

\end{abstract}

\section{Introduction}

\begin{figure*}[t]
    \centering    
    \includegraphics[width=1\textwidth]{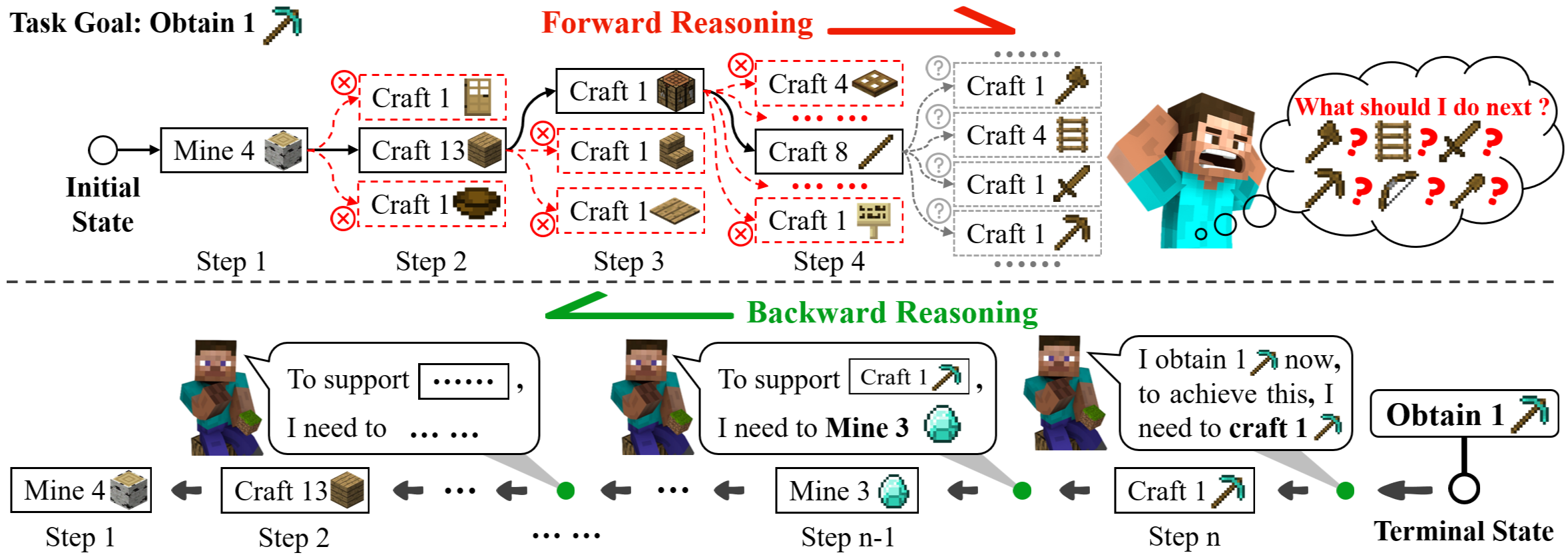}
    \caption{A toy example showing the essential advantage of backward reasoning over forward reasoning for task planning. In Minecraft, most tasks are started by ``Mine \raisebox{-1pt}{\includegraphics[height=1em]{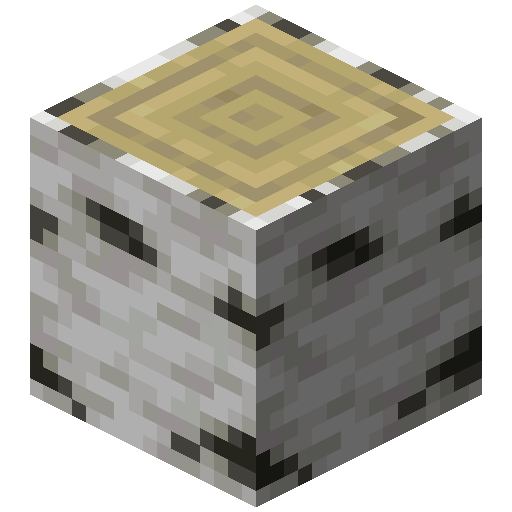}}~(log)''. 
    Best viewed in color.}
    \label{fig:forward_backward}
    \vspace{5pt}
\end{figure*}

Large language model~(LLM) based agents are a powerful tool to support human-machine intelligence because of their potential to follow human instructions. However, it is not possible to complete one complex task in a single step. The key lies in how to decompose the task into easily executed steps.
Existing LLM-based approaches mostly focus on forward reasoning based planning~\cite{huang2024understanding, koh2024tree, yu2024flow, chen-etal-2024-tree}. They infer what steps should be executed next starting from the agent's initial state, where no steps have yet been executed. {However, we observe this reasoning paradigm doesn't work well for complex tasks. Let us take a close look at a complex task~(obtain a diamond pickaxe) in an open-world environment, Minecraft, which simulates numerous complex tasks based on real-world scenarios. As shown in the upper part of Figure~\ref{fig:forward_backward}, the forward reasoning agent fails to infer the correct next step after the step ``Craft 8 \textimage{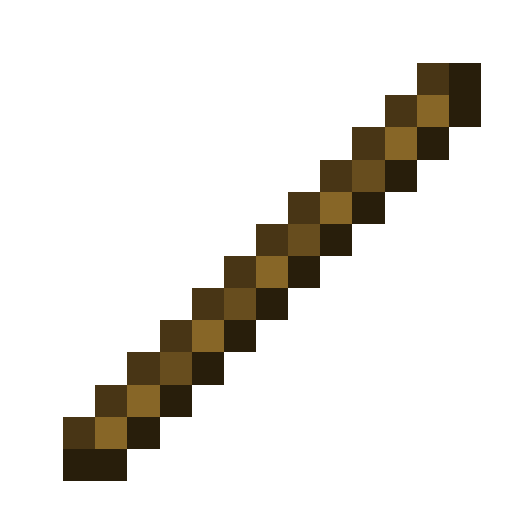}~(stick)''. This is because there is a huge perception gap between the agent's initial state and the task goal.}

\begin{figure*}[t]
    \centering    
    \includegraphics[width=1\textwidth]{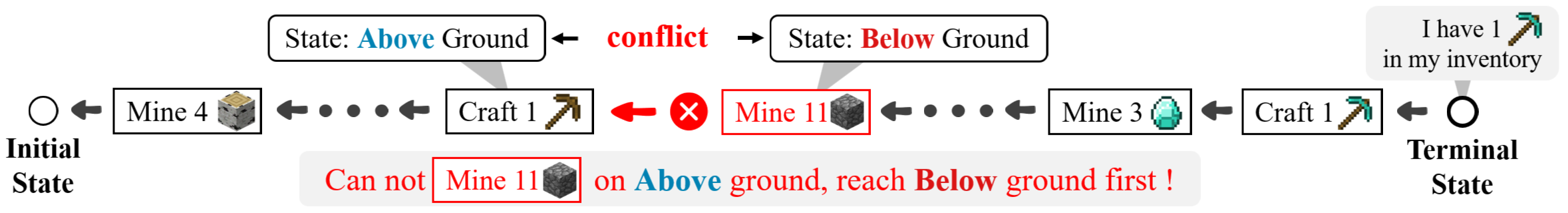}
    \caption{The conflict of the agent's physical states during planning. In Minecraft, \raisebox{-1pt}{\includegraphics[height=1em]{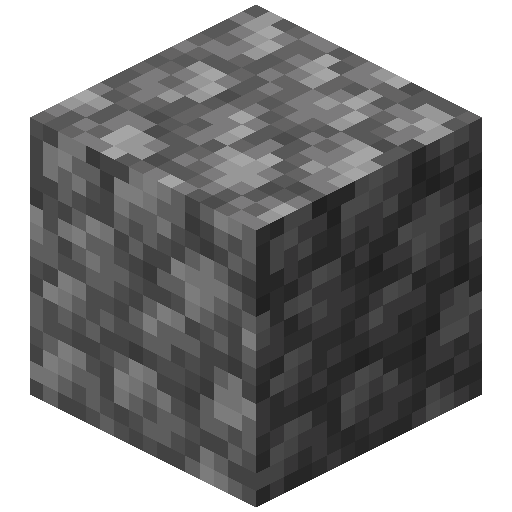}}~(stone) can only be obtained below the ground, so it is necessary to reach below the ground first and then mine \raisebox{-1pt}{\includegraphics[height=1em]{Icons/cobblestone.png}}~(stone). Best viewed on screen.}
    \label{fig:defect_of_backward}
    \vspace{5pt}
\end{figure*}


To tackle this issue, we propose to use backward reasoning~\cite{yu2023natural}, another powerful reasoning paradigm\footnote{In fact, planning by reasoning backwards has a long history outside of the LLM community, dating back to  \citet{newell1959report}, who articulated their ``principle of subgoal reduction'' in their General Problem Solver (GPS); subsequently ''regression planning'' became a major focus in the symbolic planning community (e.g. \cite{mcdermott1991regression}).}. 
As shown in the lower part of Figure~\ref{fig:forward_backward}, by backward reasoning, the agent infers the steps to be executed starting from the terminal state. In the terminal state, the task goal has been achieved and the agent has already obtained 1 \textimage{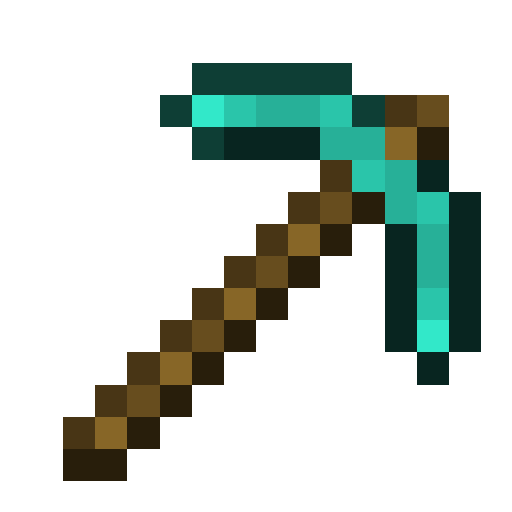}~(diamond pickaxe). To achieve this goal, it is easy to infer that the last step should be ``Craft 1 \textimage{Icons/diamond_pickaxe.png}'' because after executing this step the agent can obtain 1 diamond pickaxe. Furthermore, to successfully execute the step ``Craft 1 \textimage{Icons/diamond_pickaxe.png}'', the agent needs to prepare the materials for it. So the second to last step should be ``Mine 3 \textimage{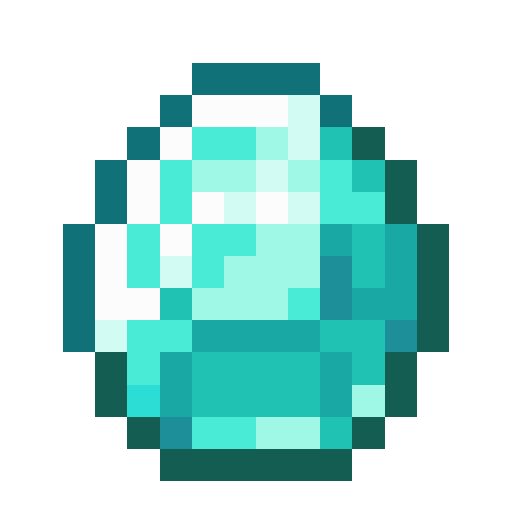}~(diamond)'' as \textimage{Icons/diamond.png} is the key material for crafting a diamond pickaxe. Subsequently, the agent can continue to infer the third to last step to support the execution of step ``Mine 3 \textimage{Icons/diamond.png}'', repeating until the entire plan is generated. \par
Despite this advantage,  backward reasoning is not  a perfect solution --- the agent only considers the current goal to be achieved, ignoring the current world state. 
As shown in Figure~\ref{fig:defect_of_backward}, the agent infers the step ``Mine \textimage{Icons/cobblestone.png}~(stone)'' immediately after inserting the step ``Craft 1 \textimage{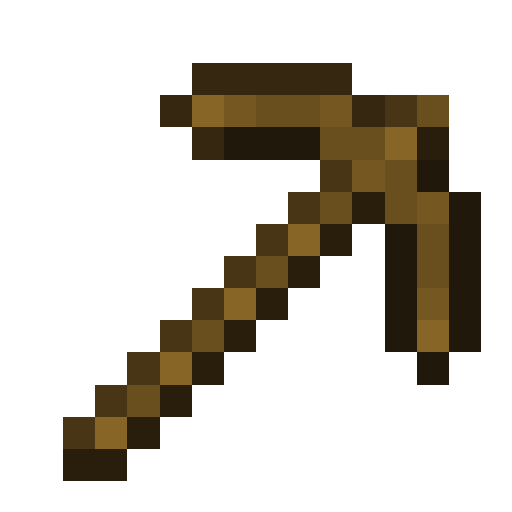}~(wooden pickaxe)''. However, this step cannot be executed successfully because \textimage{Icons/cobblestone.png}~(stone) can only be obtained below the ground but the agent was above the ground before executing this step. At last, this conflict results in a failing planning for the task.\par

To this end, we propose \model, a \underline{BA}ckward \underline{R}easoning based agent for complex Minecraft tasks, which is equipped with three carefully-designed modules. (1) Inspired by the dynamic programming algorithm~\cite{bellman1954theory}, we 
design the recursive goal decomposition module to support  robust backward planning starting from the terminal state. (2) We develop the state consistency maintaining module to monitor the agent's states and handle any state conflicts during planning. (3) We introduce the stage memory module to make the best of existing interactions and promote more efficient planning.
Experimental results show that the \model agent outperforms other SOTA methods by a large margin.
We hope our exploration and discoveries can promote the development of more efficient LLM-based planning agents.
In summary, our contributions are three-fold:
\begin{itemize}[nosep]
    \item We emphasize backward reasoning over forward reasoning for complex tasks, as it can avoid the issue caused by the perception gap between the agent's initial state and the task goal.
    
    \item We propose \model and make robust, consistent, and efficient planning from the recursive goal decomposition, state consistency maintaining, and stage memory modules.  

    \item Experimental results in Minecraft demonstrate the superiority of \model and the effectiveness of the proposed modules. 
\end{itemize}

\section{Related Work}

\subsection{Task Planning by Agent}

Task planning aims at decomposing  tasks into steps which can be directly executed. 
Current LLM-based approaches mostly focus on forward reasoning based planning. Some studies conduct iterative forward reasoning to generate the steps to be executed starting from the agent's initial state~\cite{sanyal2022fairr, huang2022language, xue2023rcot, jiang2024forward, koh2024tree, yu2024flow}. However, these studies fail to generate correct plans due to the big perception gap between the agent's initial state and task goal. 
To reduce the difficulty of planning, some studies choose the best steps from a set of pre-defined steps and organize them to form the plans for the tasks~\cite{dalvi2021explaining, creswell2022faithful, qu2022interpretable, hong2022metgen, creswell2022selection, kazemi2023lambada, zhang2024reverse}. However, these studies are limited by the pre-defined steps set and customized for specific tasks, resulting in poor generalization ability. 
Although some studies have made initial attempts to conduct planning from back to front~\cite{li2024optimus, baai2023plan4mc, wangvoyager}, they require a significant amount of trial and error before acquiring sufficient knowledge for planning retrieval~(e.g., Optimus-1~\cite{li2024optimus} accumulates historical experience through iterative interaction with the environment to support plan retrieval from the knowledge graph). 

To alleviate the above defects, we design a backward reasoning based agent that is free from the impact of the big perception gap between the agent's initial state and task goal, and enable it to handle various tasks by decomposing the task goals into natural language steps and sub-goals with LLM. Moreover, our agent supports direct static planning for any complex task by recursive goal decomposition, achieving good results even without interacting with the environment. 

\subsection{Agent for Minecraft}
Minecraft provides a simulation environment with a high degree of freedom and a wide range of tasks that can be decomposed and executed, making it an ideal resource for exploring the planning ability of LLM-based agents. 
Some studies rely on extensive interaction with the environment to gradually piece together the proper plans~\cite{wangvoyager, zhu2023ghost, wang2023jarvis, liu2024rl}. Other studies rely on a pre-designed capabilities library to generate the plans using human-written rules~\cite{yuan2023skill, wang2023describe, zhao2023see}. 
However, these studies require a significant amount of effort to trial and error and only support limited tasks with specific customization. Thus the trained models have high coupling with specific tasks and cannot generalize to other tasks. 
To address these defects, we first enable our agent to support static planning with LLMs that do not require extra customization for particular tasks, and then equip it with the ability to improve the efficiency of planning by utilizing the agent's interaction with the environment. 


\section{Method}
\label{sec:method}

\begin{figure*}
    \centering
    \includegraphics[width=1\textwidth]{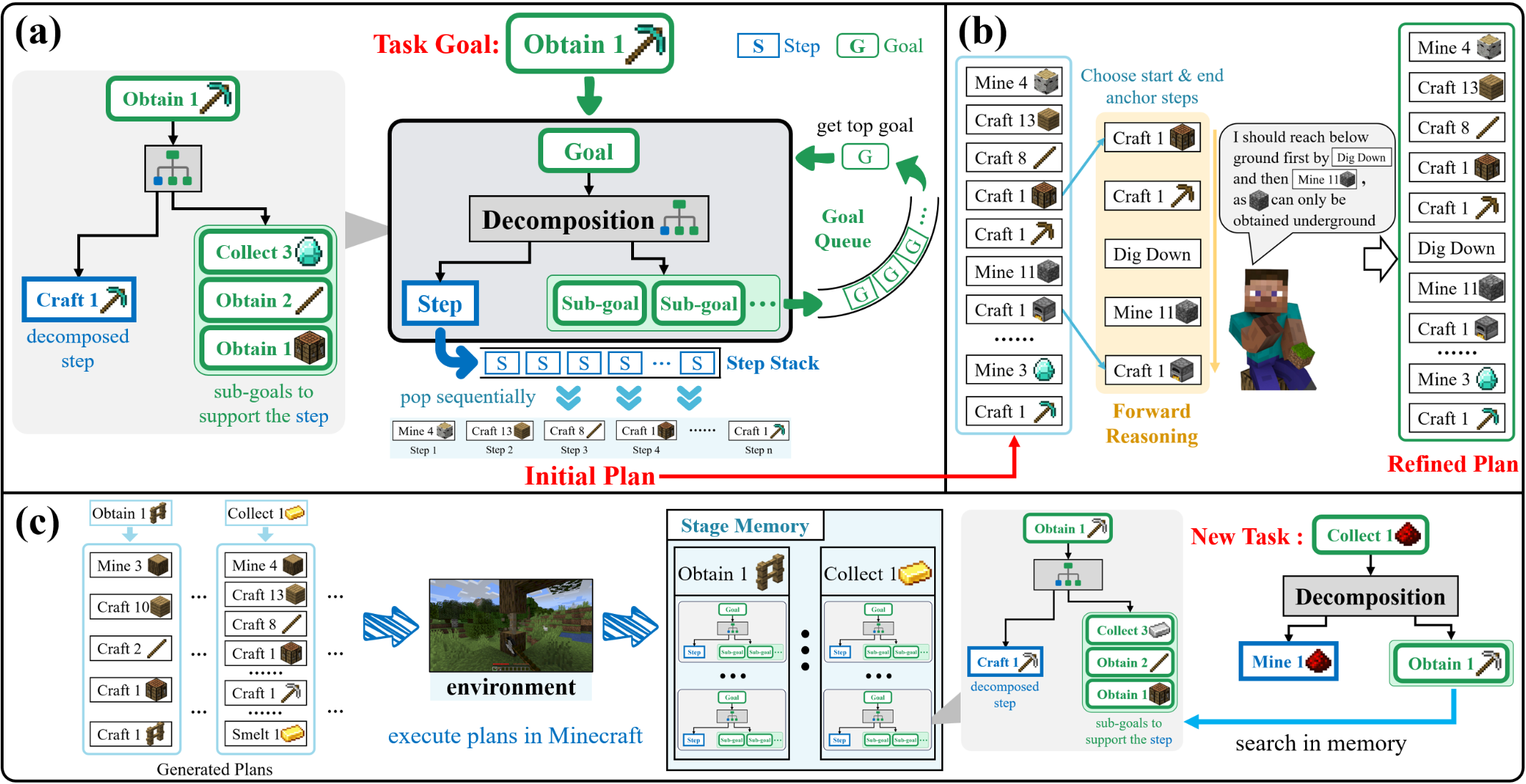}
    \caption{Our proposed backward reasoning based agent. The agent consists of three key modules. Module (a) is used to generate robust plans through recursive goals decomposition. Module (b) is used to eliminate the state conflicts in different steps by integrating forward and backward reasoning. Module (c) is used to further improve the efficiency of planning by interacting with the environment. Best viewed on screen.}
    \label{fig:method}
\end{figure*}


Our agent \model consists of three key modules: (a) recursive goal decomposition module; (b) state consistency maintaining module; (c) stage memory module. The first two modules are designed to provide robust and consistent planning, and the third module is designed to further improve planning efficiency by interacting with the environment. 

\subsection{Recursive Goal Decomposition}
\label{sec:decomposition}
To tackle the challenge brought by the big perception gap between the agent's initial state and task goal, we propose to conduct the planning starting from the terminal state (or goal state). Moreover, since decomposing one task into steps in a single iteration is extremely challenging, we propose to decompose each task through multiple iterations. Drawing inspiration from dynamic programming~\cite{bellman1954theory}, we introduce sub-goals as the transition between the task goal and the final generated plan to alleviate this planning difficulty in a recursive manner. 


In the dynamic programming algorithm~\cite{bellman1954theory}, the solution to the current problem in each iteration is composed of two parts —— (1) the immediately obtainable result and (2) the remaining results that need further solving in subsequent iterations. Inspired by this, we design a recursive goal decomposition module to split each task goal into one step and some sub-goals that need further decomposition. 
As shown in Figure~\ref{fig:method}~(a), given the task goal ``Obtain 1 \textimage{Icons/diamond_pickaxe.png}~(diamond pickaxe)'', in the first iteration we decompose the goal into two parts: (1) a step needs to be executed to achieve the goal~(``Craft 1 \textimage{Icons/diamond_pickaxe.png}'') and (2) sub-goals that are necessary to support the execution of the decomposed step~(``Collect 3 \textimage{Icons/diamond.png}'', ``Obtain 2 \textimage{Icons/stick.png}'', and ``Obtain 1 \textimage{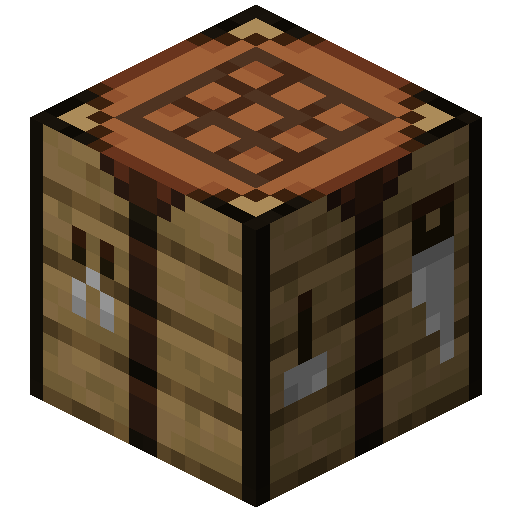}''). 

After finishing one iteration of such a decomposition, the decomposed step is pushed into a ``\textcolor{Step}{Step Stack}'' that stores all the steps to form the final plan, and the decomposed sub-goals are sequentially queued into a ``\textcolor{Goal}{Goal Queue}'' that stores the remaining goals that need to be decomposed in subsequent iterations. In the next iteration, we fetch a new goal from the head of the \textcolor{Goal}{Goal Queue} and repeat the above decomposition process to the new goal. We continuously perform such decompositions until the \textcolor{Goal}{Goal Queue} is empty. Finally, we sequentially pop the steps from the \textcolor{Step}{Step Stack} to obtain the final plan for the given task. Since there may be multiple steps of the same type, we fuse them into a single step by adding up the numbers of the items in these steps, for example, fuse ``Obtain 3 \textimage{Icons/stick.png}'' and ``Obtain 2 \textimage{Icons/stick.png}'' into ``Obtain 5 \textimage{Icons/stick.png}''. The pseudocode of the recursive goal decomposition module is shown in Algorithm~\ref{algorithm:decompose}. 

\begin{algorithm}[!ht]
	\renewcommand{\algorithmicrequire}{\textbf{Input:}}
	\renewcommand{\algorithmicensure}{\textbf{Output:}}
	\caption{Recursive Goal Decomposition}
	\label{algorithm:decompose}
 
	\begin{algorithmic}[1]
            \REQUIRE Task goal $G$
            \STATE Initialization: goal\_queue $GQ$ = $[G]$, step\_stack $SS$ = $[]$, plan $P$ = $[]$

		\REPEAT
		\STATE top\_goal $\leftarrow$ $GQ$.get\_top\_goal()
		\STATE decomposed\_step, sub\_goals $\leftarrow$ Decompose(top\_goal)
		\STATE $SS$.push(decomposed\_step)
            \FOR{$i$ in range (len(sub\_goals)) }
            \STATE $GQ$.push(sub\_goals[$i$])
            \ENDFOR
		\UNTIL $GQ$ is empty

            \REPEAT
            \STATE top\_step $\leftarrow$ $SS$.pop()
            \STATE $P$.append(top\_step) 
            \UNTIL $SS$ is empty
            
		\ENSURE $P$
	\end{algorithmic}  
\end{algorithm}


Through this module, we avoid the big perception gap between the agent's initial state and task goal by decomposing the task goal starting from the terminal state. And we reduce the difficulty of planning through multiple iterations of decomposition. Moreover, the decomposed steps and sub-goals exist in the form of natural language, without relying on external predefined steps set or customization for specific tasks. This enables our agent to handle the planning of arbitrary tasks. 

\subsection{State Consistency Maintenance}
\label{sec:consistency}
Although recursive goal decomposition is not hindered by the big perception gap between the agent's initial state and task goal, the state conflicts in different steps during planning makes it not yet a perfect solution. In fact, by forward reasoning, the agent can clearly understand its physical states before and after executing each step. Inspired by this, we propose to maintain state consistency during planning by integrating forward and backward reasoning. 

To achieve this, after generating the initial plan by module~(a), we choose pairs of start and end anchor steps from the initial plan, between which an incorrect plan may be generated due to the state conflicts between different steps. Then we adopt forward reasoning based planning to generate the partial plan between the chosen start and end anchor steps. Finally, we integrate the initial plan and the complementary partial plan to correct possible mistakes in the initial plan. 
We design two methods to choose the proper start and end anchor steps: (\Rmnum{1}) step scoring: we rate all the steps in the initial plan on a scale of 1 to 10, with 1 indicating the next few steps starting from this step are likely to be wrong and 10 indicating the next few steps starting from this step are completely accurate. Then we choose the steps with a rating below the threshold $t$ as the start anchor steps and choose the $k$-th step after each start anchor step as the end anchor steps. (\Rmnum{2}) sliding window: we randomly select the start and end anchor steps with an interval of $k$ steps from the initial plan. 

As shown in Figure~\ref{fig:method}~(b), given the initial plan of task ``Obtain 1 \textimage{Icons/diamond_pickaxe.png}'', by the step scoring method, we first rate all the steps and find that the rating of step ``Craft 1 \textimage{Icons/crafting_table.png}~(crafting table)'' is below the threshold $t$. So we choose it as the start anchor step and choose the $k$-th step after it~(``Craft 1 \textimage{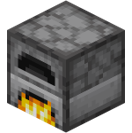}~(furnace)'') as the end anchor step. Then we adopt forward reasoning based planning to generate a partial plan between them. 
After integrating the initial plan and the generated partial plan, the agent realizes that the initial plan ignores one important step: before mining \textimage{Icons/cobblestone.png}~(stone), it should reach below ground by digging down where \textimage{Icons/cobblestone.png} exist. Therefore, the agent adds the missing step and obtains a refined plan. 

Through this module, we force the agent to understand its physical states before and after executing each step and eliminate  state conflicts that might occur between  different steps during planning. 

\subsection{Stage Memory}
\label{sec:memory}
In addition to utilizing the reasoning ability of LLMs, taking advantage of the interaction with the environment can further improve the planning efficiency of our agent. 
Current studies utilize the execution records of generated plans to support the planning of other more complex tasks~\cite{zhu2023ghost, wang2023jarvis, lin2023agentsims}. However, these studies record the execution results of entire plans that are not applicable to new tasks with different task goals, resulting in inefficient memory utilization. 

We solve this problem by designing a stage memory module. 
As our agent decomposes the task goal in multiple iterations, the decomposition result of each iteration can be used to guide subsequent goal decomposition for new tasks. To pick out accurate decomposition results, we execute the plans generated by modules (a) and (b) in the environment. The higher the execution success rate, the more accurate the decomposition result of the corresponding plan is. 
After executing the generated plans, we record each plan's execution success rate and corresponding goal decomposition results during the planning, called stage memory. 
When planning for new tasks, we retrieve the decomposition results of the same goals from stage memory with a high execution success rate and assist in decomposing the goals in new tasks. 

As shown in Figure~\ref{fig:method}~(c), after executing generated plans of task ``Obtain 1 \textimage{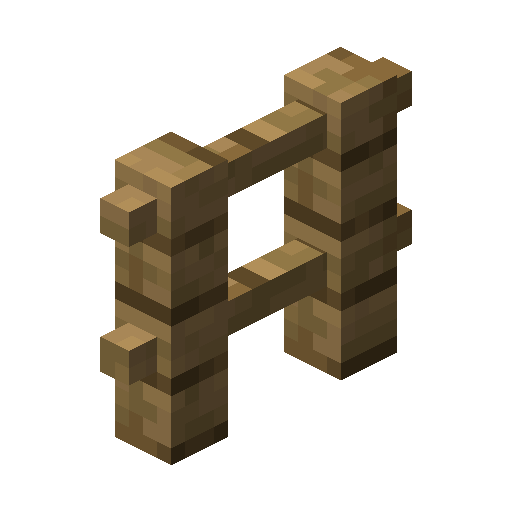}~(fence)'', ``Collect 1 \textimage{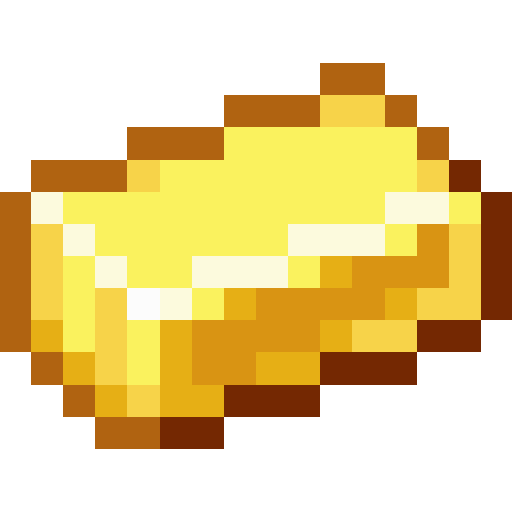}~(gold ingot)'' and so on, we record each plan's goal decomposition results and execution success rate into the stage memory. When planning for a new task ``Collect 1 \textimage{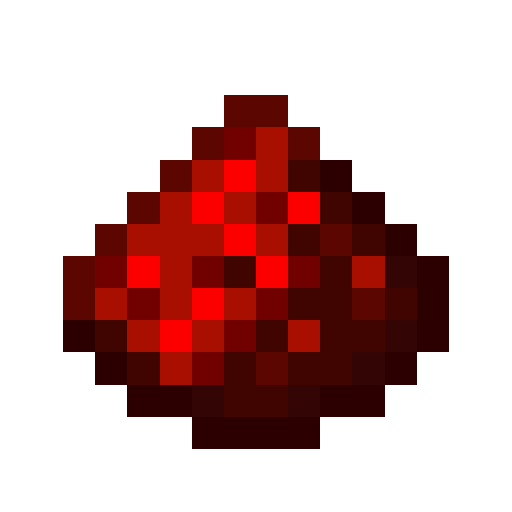}~(redstone)'', we can retrieve the decomposition result of the same goal ``Obtain 1 \textimage{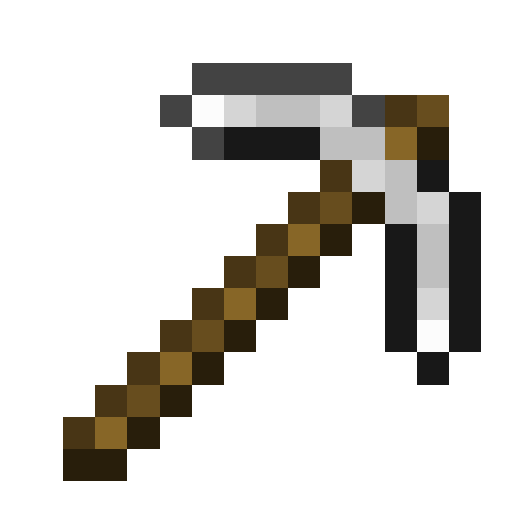}~(iron pickaxe)'' from stage memory and assist in decomposing the goal for this new task. Therefore, our agent can continuously improve the efficiency of goal decomposition with the help of stage memory. See appendix~\ref{app:our_method} for more details of our agent.

\begin{figure}[ht]
    \centering
    \includegraphics[width=.45\textwidth]{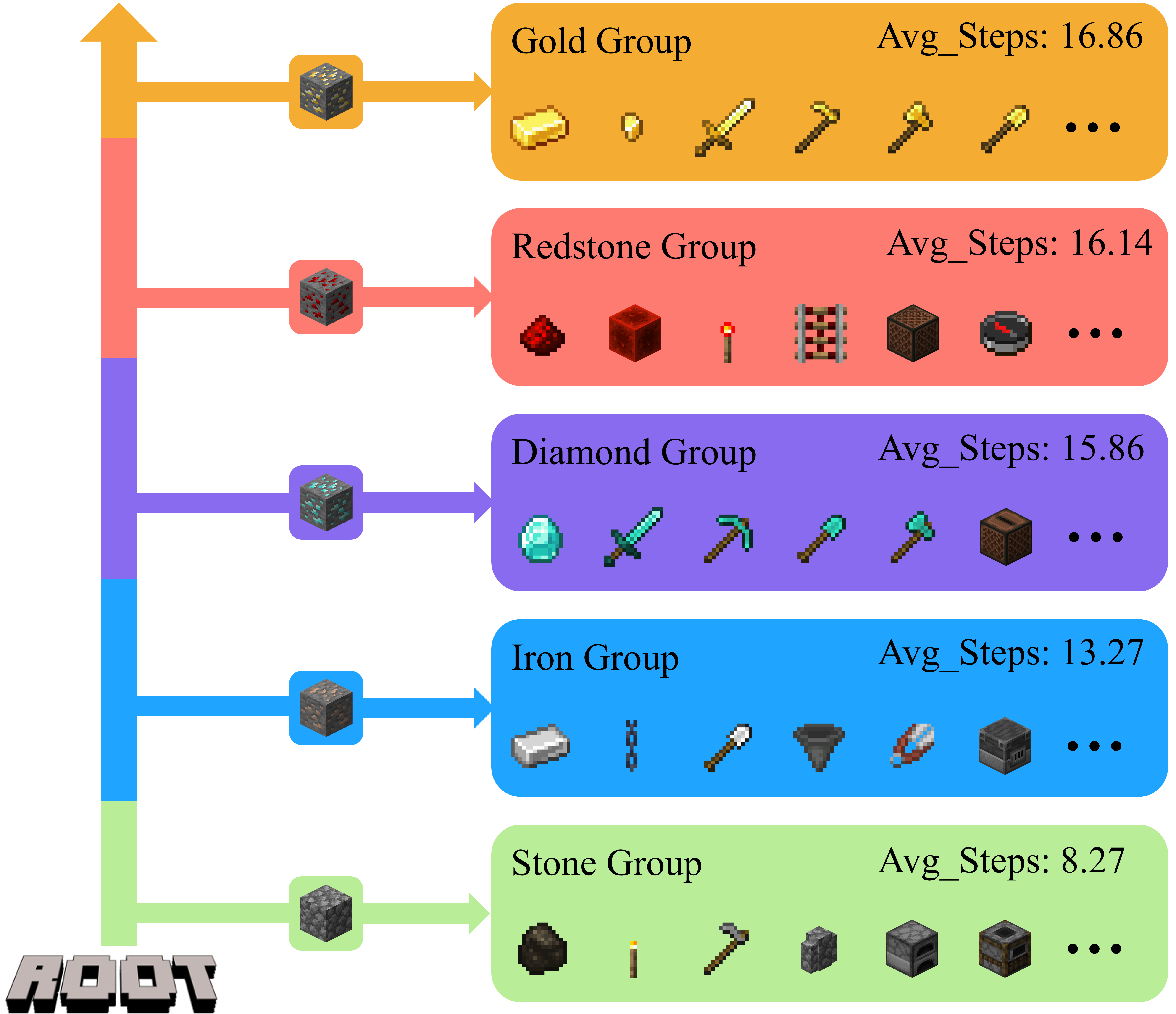}
    \caption{Collected tasks on the technology tree in Minecraft. As the depth increases, the length of the plans for the tasks gradually increases.}
    \label{fig:groups}
    \vspace{-10pt}
\end{figure}

\section{Experiments}
To evaluate the effectiveness of our proposed agent \model, we conduct extensive experiments to compare the planning performance of \model with current SOTA baselines, on both static and dynamic planning settings. 

\subsection{Experimental Setup}
\label{sec:experimental_setup}
\textbf{Static Planning Setting:}
To compare the performance of different methods, we first conduct static planning experiments with textual metrics. 
We constructed a dataset consisting of various tasks on the inherent technology tree in Minecraft. In total, we collect 53 tasks across 5 distinct groups, ranging from stone group to gold group with varying plan lengths, as shown in Figure~\ref{fig:groups}. 
We hired three experts with extensive experience in Minecraft to annotate the unique decomposed steps for each task as the ground truth, and all annotations were required to have been agreed by all three experts. 
We only compare the plans generated by modules~(a) and (b) of our agent with other methods for static planning setting, without interacting with the environment. For choosing anchor steps in section~\ref{sec:consistency}, we set $t$ to 5 and $k$ to 3, and discuss their performances in section~\ref{sec:ablation}. 
Each task was performed 10 times by each model and the average results were recorded. The temperatures of all the  LLMs used are set to $0$ for reproducibility. Finally, we report the mean performance for each group's tasks. 

\noindent
\textbf{Dynamic Planning Setting:}
To further evaluate the potential of our agent utilizing the interaction with the environment, we conduct dynamic planning experiments. 
We first execute the plans generated by modules (a) and (b) in section~\ref{sec:method} 10 times and record their average execution success rates into stage memory. Then we conduct planning again for all tasks with the help of the decomposition results corresponding to the plans with success rate $ \geqslant 0.3$ in stage memory. We report the average results of each group's tasks. See appendix~\ref{app:setting_detail} for more detailed experimental settings. 

\subsection{Static Planning}
\label{exp:static}

\begin{table*}[htbp]

\scalebox{0.7}{
\begin{tabular}{ll|ccc|ccc|ccc|ccc|ccc}
\hline
\multicolumn{2}{c|}{\multirow{2}{*}{Method}}                                 & \multicolumn{3}{c|}{\textbf{stone}}                                & \multicolumn{3}{c|}{\textbf{iron}}                                  & \multicolumn{3}{c|}{\textbf{diamond}}                               & \multicolumn{3}{c|}{\textbf{redstone}}                              & \multicolumn{3}{c}{\textbf{gold}}                                  \\ \cline{3-17} 
\multicolumn{2}{c|}{} 
& Acc $\uparrow$  & F1 $\uparrow$  & ED $\downarrow$
& Acc $\uparrow$  & F1 $\uparrow$  & ED $\downarrow$
& Acc $\uparrow$  & F1 $\uparrow$  & ED $\downarrow$
& Acc $\uparrow$  & F1 $\uparrow$  & ED $\downarrow$
& Acc $\uparrow$  & F1 $\uparrow$  & ED $\downarrow$                   \\ \hline
\multicolumn{2}{l|}{Chain-of-thought}              & \multicolumn{1}{l}{} & \multicolumn{1}{l}{} & \multicolumn{1}{l|}{} & \multicolumn{1}{l}{} & \multicolumn{1}{l}{} & \multicolumn{1}{l|}{} & \multicolumn{1}{l}{} & \multicolumn{1}{l}{} & \multicolumn{1}{l|}{} & \multicolumn{1}{l}{} & \multicolumn{1}{l}{} & \multicolumn{1}{l|}{} & \multicolumn{1}{l}{} & \multicolumn{1}{l}{} & \multicolumn{1}{l}{} \\
\multicolumn{2}{l|}{\quad - Llama-3-8B}                                & 68.97                & 68.13                & 2.91                  & 74.73                & 74.95                & 3.29                  & 55.21                & 51.47                & 8.29                  & 43.69                & 48.45                & 8.57                  & 41.28                & 52.75                & 9.00                 \\
\multicolumn{2}{l|}{\quad - GPT4}                                     & 75.82                & 75.00                & 2.09                  & 77.26                & 78.12                & 2.86                  & 68.47                & 71.94                & 4.29                  & 47.66                & 54.88                & 7.29                  & 52.54                & 73.64                & 5.86                 \\ \specialrule{0em}{1pt}{1pt} \hline

\multicolumn{2}{l|}{Reverse Chain}             & \multicolumn{1}{l}{} & \multicolumn{1}{l}{} & \multicolumn{1}{l|}{} & \multicolumn{1}{l}{} & \multicolumn{1}{l}{} & \multicolumn{1}{l|}{} & \multicolumn{1}{l}{} & \multicolumn{1}{l}{} & \multicolumn{1}{l|}{} & \multicolumn{1}{l}{} & \multicolumn{1}{l}{} & \multicolumn{1}{l|}{} & \multicolumn{1}{l}{} & \multicolumn{1}{l}{} & \multicolumn{1}{l}{} \\
\multicolumn{2}{l|}{\quad - Llama-3-8B}                                & 22.50                & 44.18                & 5.36                  & 53.28                & 65.01                & 4.67                  & 29.47                & 35.62                & 9.57                  & 13.64                & 20.84                & 12.43                 & 23.81                & 21.23                & 12.86                \\
\multicolumn{2}{l|}{\quad - GPT4}                    & 73.33                & 74.03                & 2.18                  & {\ul 80.87}          & 81.77                & {\ul 2.43}            & {\ul 86.49}          & {\ul 86.49}          & {\ul 2.14}            & {\ul 77.88}          & {\ul 77.88}          & {\ul 3.57}            & 52.17                & 74.56                & 5.43                 \\ \specialrule{0em}{1pt}{1pt} \hline

\multicolumn{2}{l|}{Self-Refine}             & \multicolumn{1}{l}{} & \multicolumn{1}{l}{} & \multicolumn{1}{l|}{} & \multicolumn{1}{l}{} & \multicolumn{1}{l}{} & \multicolumn{1}{l|}{} & \multicolumn{1}{l}{} & \multicolumn{1}{l}{} & \multicolumn{1}{l|}{} & \multicolumn{1}{l}{} & \multicolumn{1}{l}{} & \multicolumn{1}{l|}{} & \multicolumn{1}{l}{} & \multicolumn{1}{l}{} & \multicolumn{1}{l}{} \\
\multicolumn{2}{l|}{\quad - Llama-3-8B}              & 65.85                & 63.70                & 3.18                  & 69.23                & 69.58                & 4.05                  & 40.62                & 48.40                & 10.29                 & 40.57                & 47.70                & 8.57                  & 37.27                & 48.92                & 9.57                 \\
\multicolumn{2}{l|}{\quad - GPT4}                   & 75.82                & 75.00                & 2.09                  & 74.01                & 77.80                & 3.10                  & 74.77                & 79.18                & 3.29                  & 56.88                & 66.64                & 5.71                  & 52.54                & 74.09                & 5.71                 \\ \specialrule{0em}{1pt}{1pt} \hline

\multicolumn{2}{l|}{Tree-of-thought}                     & 46.91                & 43.92                & 4.27                  & 33.96                & 44.46                & 6.81                  & 38.64                & 28.43                & 10.00                 & 20.43                & 22.23                & 12.14                 & 35.71                & 30.95                & 10.14                \\
\multicolumn{2}{l|}{DEPS}                    & 49.45                & 46.20                & 3.73                  & 61.01                & 50.35                & 5.24                  & 49.55                & 41.32                & 7.86                  & 26.44                & 25.95                & 11.00                 & 48.65                & 39.42                & 8.14                 \\
\multicolumn{2}{l|}{Plan-and-Solve}          & 45.45                & 58.29                & 3.45                  & 76.17                & 80.29                & 2.71                  & 66.67                & 77.02                & 4.14                  & 42.06                & 50.67                & 8.43                  & 63.00                & 59.93                & 6.43                 \\
\multicolumn{2}{l|}{Openai-o1}                     & 74.73                & 74.86                & 2.09                  & 77.98                & {\ul 82.34}          & 2.67                  & 63.96                & 79.55                & 3.57                  & 57.52                & 76.44                & 3.86                  & {\ul 75.42}          & {\ul 83.97}          & {\ul 3.14}           \\ \specialrule{0em}{1pt}{1pt} \hline
\multicolumn{2}{l|}{Ours}                   & \multicolumn{1}{l}{} & \multicolumn{1}{l}{} & \multicolumn{1}{l|}{} & \multicolumn{1}{l}{} & \multicolumn{1}{l}{} & \multicolumn{1}{l|}{} & \multicolumn{1}{l}{} & \multicolumn{1}{l}{} & \multicolumn{1}{l|}{} & \multicolumn{1}{l}{} & \multicolumn{1}{l}{} & \multicolumn{1}{l|}{} & \multicolumn{1}{l}{} & \multicolumn{1}{l}{} & \multicolumn{1}{l}{} \\
\multicolumn{2}{l|}{\quad - Llama-3-8B}              & {\ul 78.02}          & {\ul 78.02}          & {\ul 1.82}            & 74.44                & 76.00                & 3.33                  & 72.97                & 79.67                & 3.29                  & 59.22                & 59.10                & 7.00                  & 60.71                & 66.38                & 5.86                 \\
\multicolumn{2}{l|}{\quad - GPT4}                    & {\color[HTML]{000000} \textbf{87.91}} & {\color[HTML]{000000} \textbf{87.91}} & {\color[HTML]{000000} \textbf{1.00}} & {\color[HTML]{000000} \textbf{82.67}} & {\color[HTML]{000000} \textbf{85.02}} & {\color[HTML]{000000} \textbf{2.19}} & {\color[HTML]{000000} \textbf{87.39}} & {\color[HTML]{000000} \textbf{88.23}} & {\color[HTML]{000000} \textbf{1.86}} & {\color[HTML]{000000} \textbf{80.53}} & {\color[HTML]{000000} \textbf{82.22}} & {\color[HTML]{000000} \textbf{2.86}} & {\color[HTML]{000000} \textbf{85.59}} & {\color[HTML]{000000} \textbf{87.34}} & {\color[HTML]{000000} \textbf{2.14}} \\ \hline
\end{tabular}
}
\caption{Results of the static planning experiments. Best results are highlighted in {\color[HTML]{000000} \textbf{bold}}, and the second best results are highlighted with {\ul underline}. For the adopted metrics, ``Acc'' refers to Accuracy, ``F1'' refers to F1-Score, and ``ED'' refers to Edit Distance. Larger Accuracy and F1-Score and smaller Edit Distance represent better performance.}
\label{tab:static}
\end{table*}

\textbf{Baselines:} 
(1) \textbf{Chain-of-thought}~\cite{wei2022chain}: utilize forward reasoning based thought chains to generate the plans step by step. 
(2) \textbf{Reverse Chain}~\cite{zhang2024reverse}: utilize backward reasoning based thought chains to generate the plans. 
(3) \textbf{Self-Refine}~\cite{madaan2024self}: utilize feedback from the model itself to refine the generated plans. 
(4) \textbf{Tree-of-thought}~\cite{yao2024tree}: enable the exploration of multiple forward reasoning paths and choose the best one. 
(5) \textbf{DEPS}~\cite{wang2023describe}: enhance the accuracy of planning by integrating the description and explanation of the plans given by LLM. 
(6) \textbf{Plan-and-Solve}~\cite{wang2023plan}: utilize in-context learning to rehearse generated plans and improve the logical correctness of the plans. 
(7) \textbf{Openai-o1}~\cite{o1_2024}: perform a lot of reasoning and reflection before outputting the final results to improve the accuracy of generated plans. 
See appendix~\ref{app:baseline} for implementation details of the baselines.  

\noindent
\textbf{Metrics:}
We design three textual metrics based on overlap computation to evaluate the quality of generated plans:
(1) \textbf{Accuracy}: Evaluate whether each step in the generated plan matches exactly with the corresponding step in the ground truth at the same index. 
(2) \textbf{F1-Score}: Take into account both the precision and recall of generated plans. A step is considered correctly generated only if the pair it forms with its previous step can match the ground truth. 
(3) \textbf{Edit Distance}: Evaluate the minimum number of editing operations needed to transform the generated plan into the ground truth. Editing operations involve replacing, inserting, or deleting a single step in the generated plan. See appendix~\ref{app:metric} for detailed demonstration. 

\noindent
\textbf{Results:}
As shown in Table~\ref{tab:static}, our agent achieves SOTA performance on all 5 groups under all three metrics. This demonstrates that our agent can provide robust and consistent planning for various tasks. In addition, we have the following findings: 

(1) In addition to the powerful closed-source model GPT4~\cite{achiam2023gpt}, with the open-source Llama-3 model~\cite{dubey2024llama} as backbone, our agent can still significantly outperform other baselines that also adopt Llama-3 as backbone model. This shows the robustness of our agent for planning with both closed-source and open-source LLMs. 

(2) Chain-of-thought fails to achieve good performance due to the inherent challenge brought by the big perception gap between the agent's initial state and task goal. This makes it difficult for an LLM to infer correct steps, and the situation becomes worse on tasks with longer plans. 

(3) Backward reasoning shows its superiority by comparing the performance of Chain-of-thought and Reverse Chain. Contrary to Chain-of-thought, Reverse Chain infers the steps to be executed starting from the terminal state, free from the big perception gap between the agent's initial state and task goal. However, simply changing the reasoning paradigm cannot solve the defects brought by the state conflicts in different steps during planning, resulting in a decline in performance. 

\begin{table*}[!th]

\scalebox{0.75}{
\begin{tabular}{ll|ccc|ccc|ccc|ccc|ccc}
\hline
\multicolumn{2}{c|}{}                                & \multicolumn{3}{c|}{{\color[HTML]{262626} \textbf{stone}}}                                                           & \multicolumn{3}{c|}{{\color[HTML]{262626} \textbf{iron}}}                                                            & \multicolumn{3}{c|}{{\color[HTML]{262626} \textbf{diamond}}}                                                         & \multicolumn{3}{c|}{{\color[HTML]{262626} \textbf{redstone}}}                                                        & \multicolumn{3}{c}{{\color[HTML]{262626} \textbf{gold}}}                                                             \\ \cline{3-17} 
\multicolumn{2}{c|}{\multirow{-2}{*}{Method}}
& Acc $\uparrow$  & F1 $\uparrow$  & ED $\downarrow$
& Acc $\uparrow$  & F1 $\uparrow$  & ED $\downarrow$
& Acc $\uparrow$  & F1 $\uparrow$  & ED $\downarrow$
& Acc $\uparrow$  & F1 $\uparrow$  & ED $\downarrow$
& Acc $\uparrow$  & F1 $\uparrow$  & ED $\downarrow$            \\ \hline
\multicolumn{2}{l|}{{\color[HTML]{262626} React}}   & 70.33                                 & 72.53                                 & 2.18                                 & 89.17                                 & 89.17                                 & 1.43                                 & {\color[HTML]{262626} 71.17}          & {\color[HTML]{262626} 74.77}          & {\color[HTML]{262626} 4.57}          & 57.01                                 & 57.49                                 & 7.00                                 & 57.63                                 & 79.75                                 & 4.43                                 \\
\multicolumn{2}{l|}{{\color[HTML]{262626} Jarvis-1}} & 75.82                                 & 75.00                                 & 2.09                                 & 73.61                                 & 76.56                                 & 3.38                                 & {\color[HTML]{262626} 81.98}          & {\color[HTML]{262626} 81.98}          & {\color[HTML]{262626} 2.86}          & 70.09                                 & 71.70                                 & 5.00                                 & 52.54                                 & 74.09                                 & 5.71                                 \\ \hline
\multicolumn{2}{l|}{Ours}                            & \multicolumn{1}{l}{}                  & \multicolumn{1}{l}{}                  & \multicolumn{1}{l|}{}                & \multicolumn{1}{l}{}                  & \multicolumn{1}{l}{}                  & \multicolumn{1}{l|}{}                & \multicolumn{1}{l}{}                  & \multicolumn{1}{l}{}                  & \multicolumn{1}{l|}{}                & \multicolumn{1}{l}{}                  & \multicolumn{1}{l}{}                  & \multicolumn{1}{l}{}                 \\
\multicolumn{2}{l|}{{\color[HTML]{262626} \quad - static}}   & {\color[HTML]{262626} 87.91}          & {\color[HTML]{262626} 87.91}          & {\color[HTML]{262626} 1.00}          & {\color[HTML]{262626} 82.67}          & {\color[HTML]{262626} 85.02}          & {\color[HTML]{262626} 2.19}          & {\color[HTML]{262626} 87.39}          & {\color[HTML]{262626} 88.23}          & {\color[HTML]{262626} 1.86}          & {\color[HTML]{262626} 80.53}          & {\color[HTML]{262626} 82.22}          & {\color[HTML]{262626} 2.86}          & 85.59                                 & 87.34                                 & 2.14                                 \\
\multicolumn{2}{l|}{\quad - dynamic}                         & {\color[HTML]{000000} \textbf{92.31}} & {\color[HTML]{000000} \textbf{92.31}} & {\color[HTML]{000000} \textbf{0.64}} & {\color[HTML]{000000} \textbf{96.58}} & {\color[HTML]{000000} \textbf{96.92}} & {\color[HTML]{000000} \textbf{0.45}} & {\color[HTML]{000000} \textbf{88.29}} & {\color[HTML]{000000} \textbf{89.23}} & {\color[HTML]{000000} \textbf{1.71}} & {\color[HTML]{000000} \textbf{86.73}} & {\color[HTML]{000000} \textbf{86.73}} & {\color[HTML]{000000} \textbf{2.14}} & {\color[HTML]{000000} \textbf{86.44}} & {\color[HTML]{000000} \textbf{94.93}} & {\color[HTML]{000000} \textbf{0.86}} \\ \hline
\end{tabular}
}
\caption{Results of the dynamic planning experiments by textual metrics. Best results are highlighted in {\color[HTML]{000000} \textbf{bold}}. For the adopted metrics, ``Acc'' refers to Accuracy, ``F1'' refers to F1-Score, and ``ED'' refers to Edit Distance. 
``Ours-static'' refers to our agent with only modules (a) and (b) in static planning setting, GPT4 as the backbone model.  
``Ours-dynamic'' refers to our complete agent, including the stage memory module, GPT4 as the backbone model.}
\label{tab:dynamic_textual}
\end{table*}

(4) Self-Refine and DEPS can improve the plans by obtaining feedback from the LLM. However, as the initial plans are generated based on forward reasoning, they face the same challenge as Chain-of-thought. Moreover, severe error propagation during multiple turns of refinement may introduce irrelevant noise to the generated plans, and the unstable feedback from the LLM can mislead the agent's reasoning to the wrong direction. 

(5) Although Tree-of-thought can explore multiple paths during planning, evaluating the quality of incomplete plans by LLM is unreliable. Moreover, due to the interference of a large number of candidate steps, the generated plans are mixed with unnecessary noisy steps, resulting in poor performance. 

(6) Plan-and-Solve and Openai-o1 can improve the plans by previewing the decomposed partial plans and repeatedly considering more possible solutions. However, even with more exploration of different paths and reflection on generated plans, it is still difficult for an LLM to infer correct steps that are far from the task goal. 

\subsection{Dynamic Planning}
\label{sec:dynamic}
To further evaluate the potential of our agent utilizing the interaction with the environment, with the help of our designed stage memory described in section~\ref{sec:memory}, we compare our agent with other baselines that also utilize the interaction with the environment for planning. 

\noindent
\textbf{Baselines:} 
(1) \textbf{React}~\cite{yao2023react}: combine reasoning and acting to conduct planning. 
(2) \textbf{Jarvis-1}~\cite{wang2023jarvis}: conduct planning in Minecraft with both pre-trained knowledge in LLMs and the game experiences given by the environment. 
(3) \textbf{Ours-Static}: our agent with only modules (a) and (b) in static planning setting, GPT4 as the backbone model. 
(4) \textbf{React + Ours-Static}: execute the generated plans by \textbf{Ours-Static}, but under the reasoning and acting framework of \textbf{React}. 


\noindent
\textbf{Metrics:}
In addition to the textual metrics used in the static planning experiments, we further record the success rates and computation time for executing the generated plans in the environment~( execution metrics). For a fair comparison, we execute all evaluated tasks 10 times for and report the average execution success rates and computation time for tasks in each group. 

\begin{figure}[th]
    \centering
    \includegraphics[width=.48\textwidth]{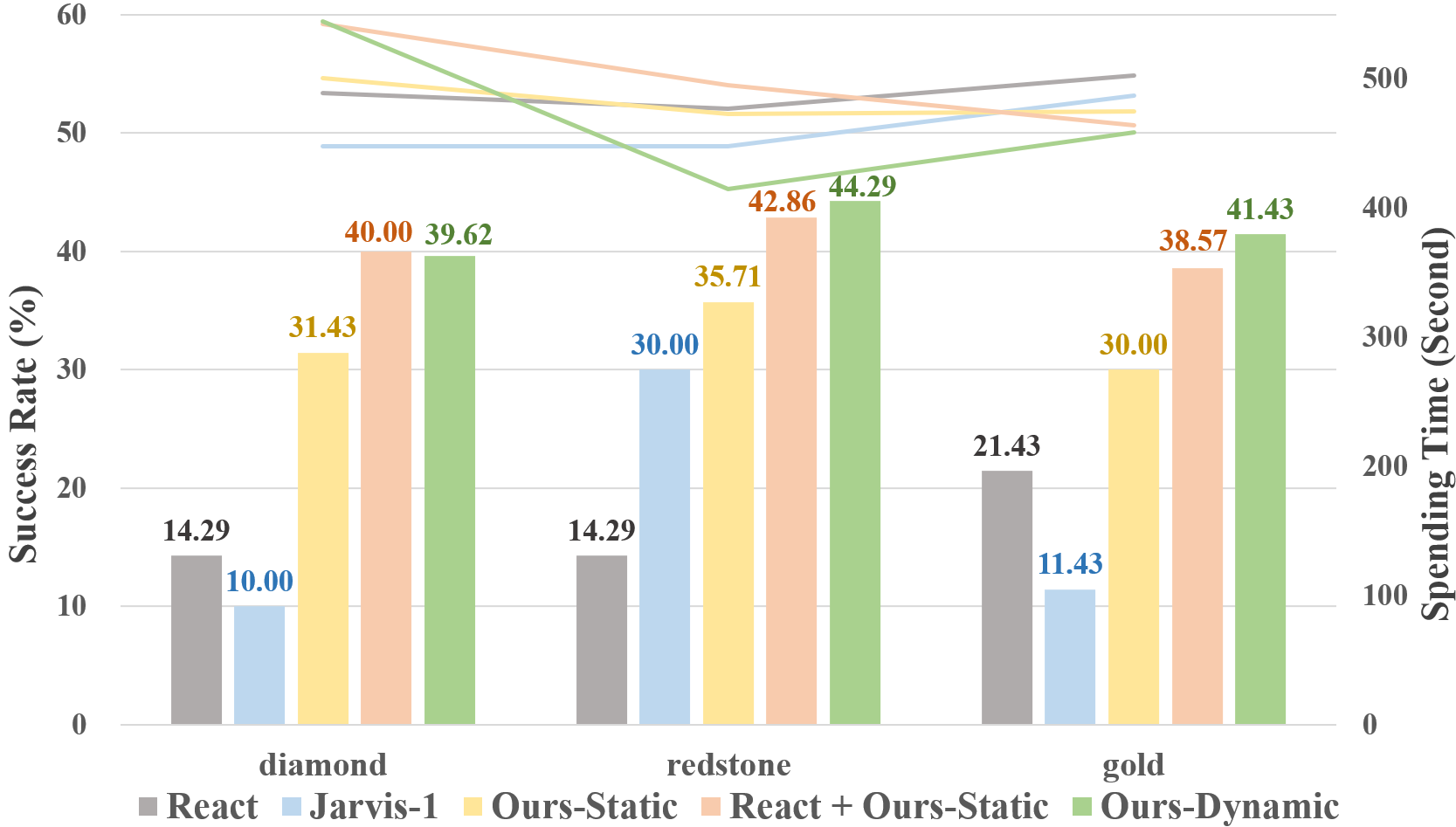}
    \caption{Results of the dynamic planning experiments by execution metrics. The bars represent the average execution success rates of tasks in different groups, and the lines represent the average time spent to execute the takes.}
    \label{fig:dynamic}
    \vspace{-12pt}
\end{figure}

\noindent
\textbf{Results:}
As shown in Table~\ref{tab:dynamic_textual} and Figure~\ref{fig:dynamic}, with the support of the stage memory module, our agent beats the SOTA performance by utilizing the interaction with the environment, both in terms of textual metrics and execution metrics. Additionally, we have the following observations: \\
(1) Based on our designed stage memory module, our agent can obtain the feedback of generated plans by interacting with the environment, thereby improving the efficiency of subsequent goals decomposition. With just one round of interaction, the quality of the generated plans can be significantly improved. 
(2) React and Jarvis-1 can achieve good performance by interacting with the environment. However, they cannot solve the big perception gap between the agent's initial state and task goal during planning. Even if the feedback from the environment can indicate the execution results of generated plans, no effective information is given to refine the incorrect plans. 
(3) Taking the plans generated by our agent as the reference, React can significantly improve the execution success rates of plans. This indicates that high-quality plans are key in helping agents successfully complete tasks in the environment. Therefore, relying solely on environmental feedback to correct the generated plans is not the most effective solution. A better approach is to focus on improving the quality of the generated initial plans by LLM. 
(4) Environmental feedback can indeed help improve the quality and execution success rate of plans. But compared with React and Jarvis-1, our agent can more effectively utilize the interaction with the environment through stage memory, as we can efficiently take advantage of every piece of decomposition results in all tasks. 

\subsection{Ablation Study}
\label{sec:ablation}

\begin{figure}[th]
    \centering
    \includegraphics[width=.45\textwidth]{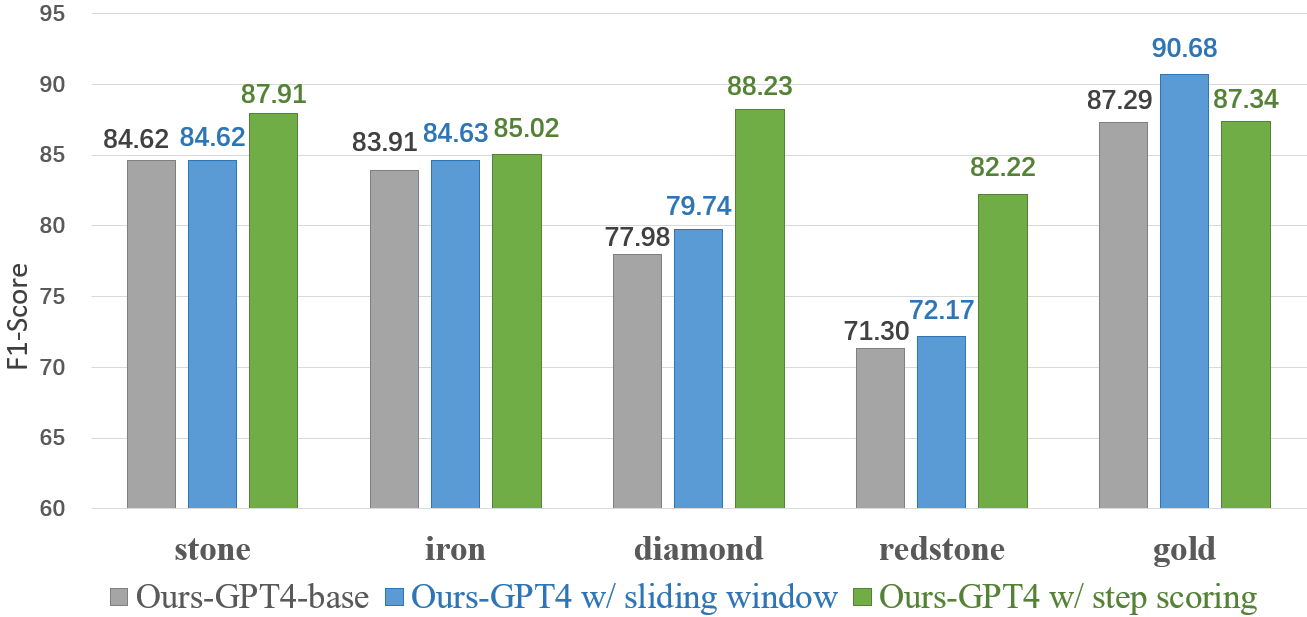}
    \caption{Results of the ablation study for state consistency maintaining module, GPT4 as the backbone model.}
    \label{fig:ablation_GPT4}
\end{figure}

\begin{figure}[th]
    \centering
    \includegraphics[width=.45\textwidth]{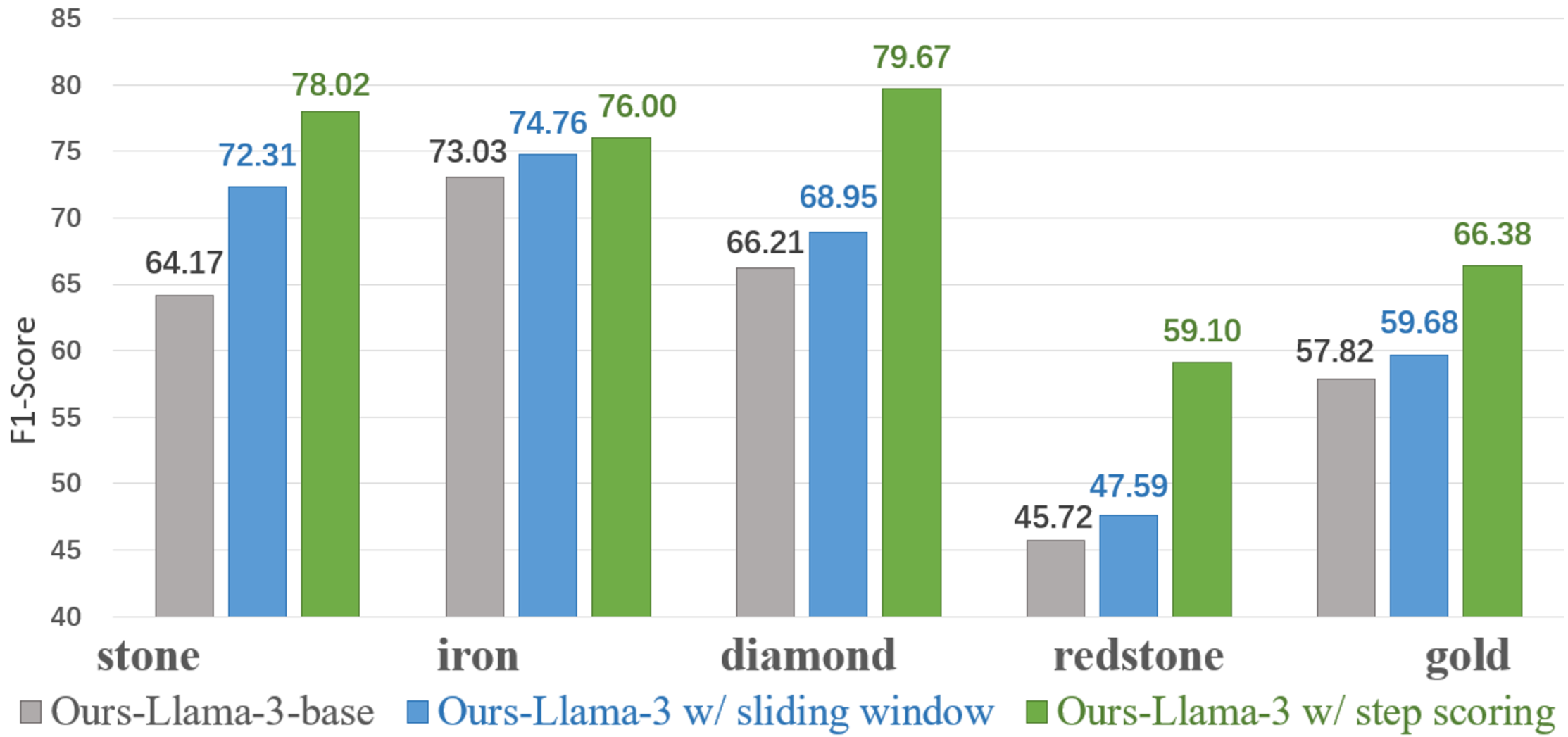}
    \caption{Results of the ablation study for state consistency maintaining module, Llama-3-8B as the backbone model.}
    \label{fig:ablation_Llama3}
\end{figure}

The effectiveness of our proposed stage memory module has been verified in section \ref{sec:dynamic}. 
Now we conduct an ablation study to verify the effectiveness of the state consistency maintaining module and compare the two methods to choose anchor steps described in section~\ref{sec:consistency}. 
As shown in Figure~\ref{fig:ablation_GPT4} and~\ref{fig:ablation_Llama3}, our proposed state consistency maintaining module can significantly improve the quality of generated plans. Whether using GPT4 or Llama-3 as the backbone model, it can effectively eliminate the state conflicts in different steps by integrating forward and backward reasoning. 
The open-source Llama-3 model can benefit more from it because the initial plans' quality of Llama-3 is lower than that of GPT4. 
Additionally, by comparing the two methods to select anchor steps, we find that step scoring can bring greater performance improvement. This is because the sliding window method chooses too many anchor step pairs and introduces unnecessary noise into the generated plans. 

\section{Conclusion}
In this paper, we study the planning problem for complex tasks in the Minecraft environment. We argue that, compared with forward reasoning, backward reasoning has the advantage to avoid the perception gap between the agent's initial state and the task goal. We further point out a key weakness of backward reasoning is that it may fail to complete the task because of the state conflicts during planning. To this end, we propose \model with: (1)   recursive goal decomposition to make robust backward reasoning, (2) a state consistency maintaining module to address  state conflicts, (3) and a stage memory module promote more efficient planning. Experimental results show the superiority of our agent in both static and dynamic planning, surpassing other baselines by a large margin. 

\section{Ethical Considerations}
The dataset we constructed does not contain any personal, private or sensitive information and is only used for research purposes. The participation of volunteer human annotators in the construction of the dataset does not constitute an ethical concern, as all annotators were informed of the nature of the task, participated voluntarily without coercion, and no sensitive personal data was collected or disclosed during the annotation process. Therefore, we believe that there is no ethical issue with our work.

\section{Limitations}
To verify the generalizability of our work, tests on many further tasks should be undertaken,
but we believe that as a primary step towards Backward Reasoning based Planning with LLMs, Minecraft is rich enough to provide many scenarios and tasks in an open-ended environment that simulates the real world. As described in section \ref{sec:experimental_setup}, we have collected a large number of tasks with various difficulty levels in different scenarios to demonstrate that our method can handle a wide range of tasks. 
Moreover, as demonstrated in prior work, Minecraft is a sufficiently complex environment to undertake a rich variety of planning experiments. We leave exploration in other environments as future work. 

\section*{Acknowledgement}
This work was supported in part by the National Natural Science Foundation of China (No. 62206191, No. 62272330, and No. U24A20328); in part by the Science Fund for Creative Research Groups of Sichuan Province Natural Science Foundation (No. 2024NSFTD0035); in part by the Natural Science Foundation of Sichuan (No. 2024YFHZ0233).

\bibliography{custom}

\clearpage
\appendix

\section{Introduction to the Simulation Environment Minecraft}
\label{app:minecraft}
Minecraft is a simulation environment with extremely high degrees of freedom and provides a large number of scenarios and tasks similar to the real world for agents to explore. In Minecraft, intelligent agents can do anything they want in different scenarios, such as collecting various materials and crafting useful tools. Therefore, exploring the behaviors of agents in Minecraft is of great significance for serving various applications in the real world. Minecraft covers diverse types of tasks that can be executed, with sufficiently complex tasks to challenge the agents' abilities. Moreover, the simulation environment is completely under our control, which facilitates us to conduct a large number of scientific experiments. Based on these advantages, we chose Minecraft as the platform to explore the automatic task completion and planning abilities of LLM based agents. 

\section{Details of Experimental Settings}
\label{app:setting_detail}

\subsection{Details of Evaluated Tasks}
\label{app:tasks_detail}
To evaluate the planning ability of different models, we constructed a dataset consisting of various tasks on the technology tree in Minecraft. 

Considering that the evaluation metrics are based on overlap computation to assess the accuracy of generated plans, we only selected tasks with unique plans when constructing the dataset. This can make sure that the ground truth for each task is unique. For a few flexible tasks, we first standardize the plans generated by the model according to our pre-defined unified standards and then compare them with the ground truth to ensure fair comparison (e.g., always ensure that the step ``craft 1 crafting\_table'' is before the step ``craft 1 stick''). 

Details of the chosen tasks for evaluation are listed in Table~\ref{tab:tasks_detail}.
\begin{table}[!htb]
\caption{Details of Evaluated Tasks}
\scalebox{0.8}{
\begin{tabular}{cccc}
\hline
\textbf{Group} & \textbf{Task Num} & \textbf{Avg Plan Length} & \textbf{Example} \\ \hline
Stone          & 11                  & 8.27                    & obtain 1 \textimage{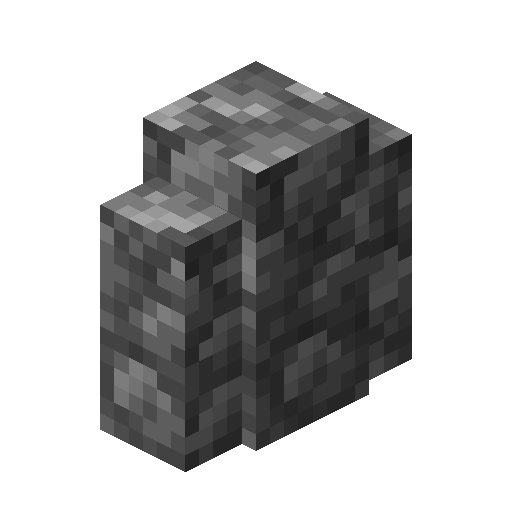}                 \\  
Iron           & 21                  & 13.27                   & obtain 1 \textimage{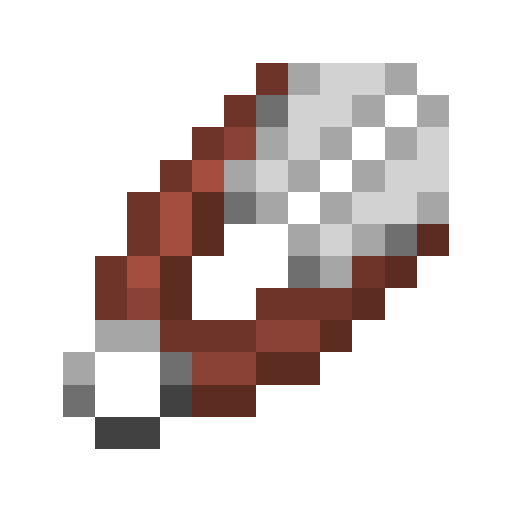}                   \\
Diamond        & 7                   & 15.86                   & obtain 1 \textimage{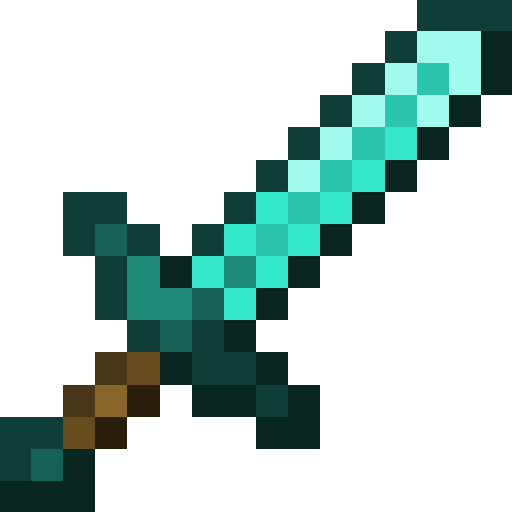}                   \\
Redstone       & 7                   & 16.14                   & obtain 1 \textimage{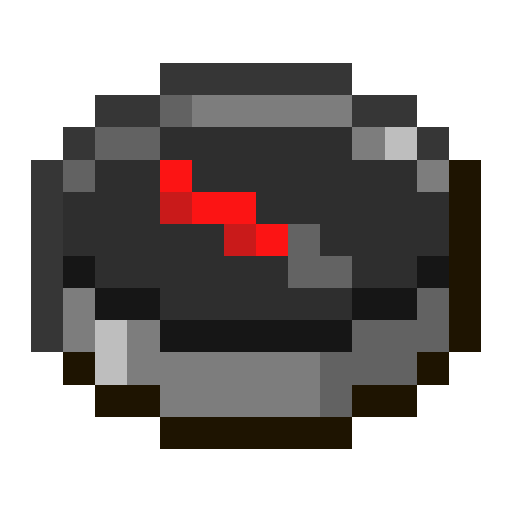}                   \\
Gold           & 7                   & 16.86                   & obtain 1 \textimage{Icons/golden\_axe.png}                 \\ \hline
\end{tabular}
}
\label{tab:tasks_detail}
\end{table}



\subsection{Details of Adopted Textual Metrics}
\label{app:metric}

\begin{figure}[th]
    \centering
    \includegraphics[width=.48\textwidth]{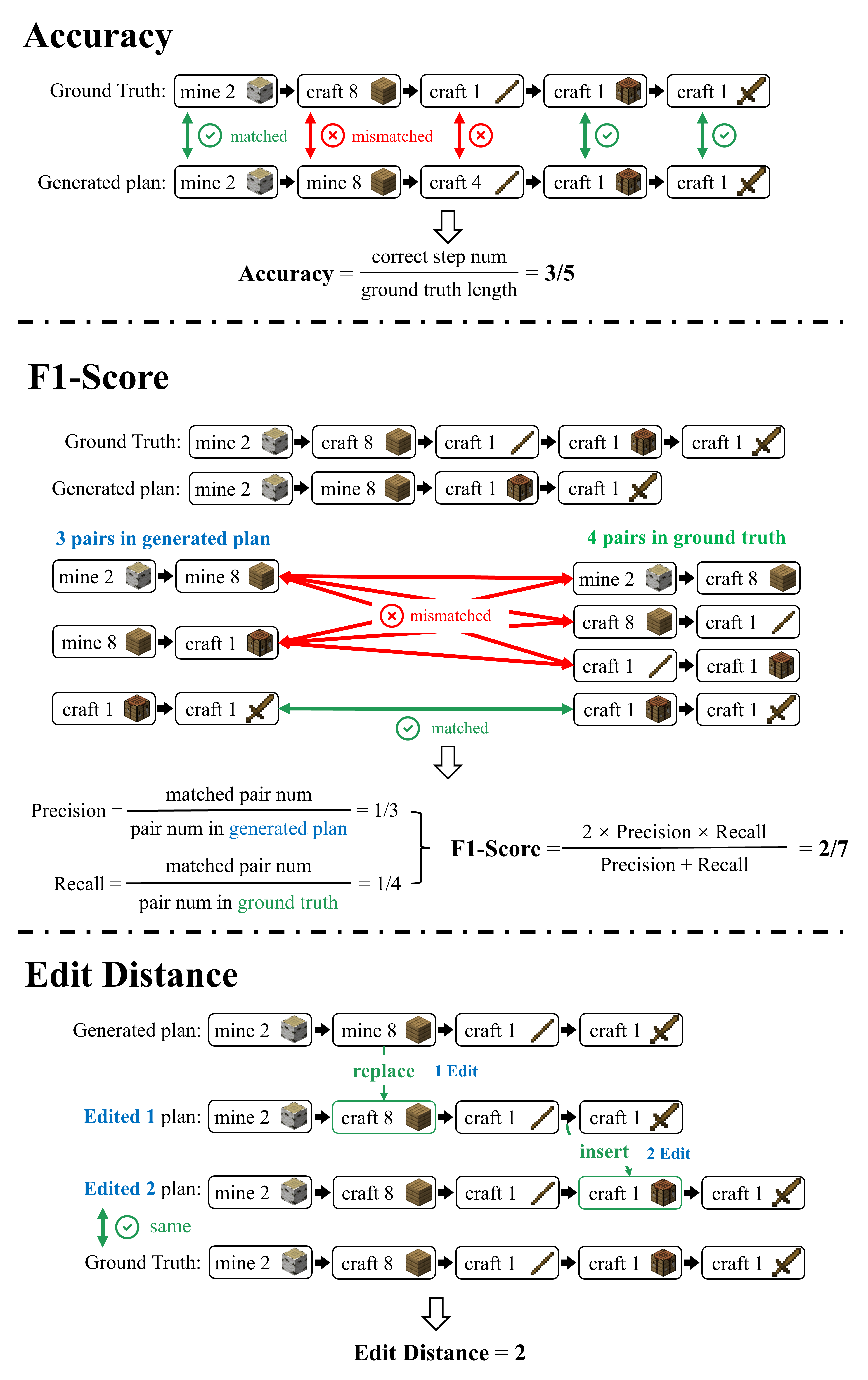}
    \caption{Computation demonstration of the three adopted textual metrics.}
    \label{fig:metric}
\end{figure}

We design three textual metrics based on overlap computation to evaluate the quality of generated plans. The computation demonstration of the three metrics is shown in Figure~\ref{fig:metric}.

\subsection{Details of the Simulation Environment}
\label{app:environment}




\subsubsection{High-level Steps Encapsulation}
\label{app:encapsulation}
As our work focuses on task planning instead of generating fine-grained machine instructions, based on the work of Jarvis-1~\cite{wang2023jarvis}, we utilize our game experience and knowledge of Minecraft to encapsulate complex operations into high-level steps that align with the steps in generated plans by our agent and baseline models. 
The action space in Minecraft includes keyboard and mouse operations that are usually performed by humans. For example, using the keyboard to control the agent's movement, jumping, opening or closing the inventory, and selecting items from the hotbar. Mouse movements are used to adjust the agent's view, including horizontal rotation and pitch. Mouse clicks are used for attacking mobs, breaking blocks, and interacting with crafting\_table and furnace. When using crafting\_table and furnace, precise item selection and manipulation are needed with fine-grained mouse movement control, which cannot be reliably accomplished by current low-level controller models. 

To address this problem and make sure the generated plans can be successfully executed in the environment, we propose to encapsulate the fine-grained keyboard and mouse operations into three types of frequently-used steps: \texttt{craft}, \texttt{smelt}, and \texttt{equip}. 
Heuristic rules are applied to the three types of steps for encapsulation. 
For another two types of important steps \texttt{mine} and \texttt{dig down}, we use the SOTA low-level controller model MineDreamer\cite{zhou2024minedreamer} to directly execute the natural language instructions. 
Details of the action space in Minecraft and the encapsulation of the above mentioned steps are listed in Table~\ref{tab:action_space} and Table ~\ref{tab:step_space}. 
\begin{table*}[!htb]
\caption{Details of the action space in Minecraft.}
\scalebox{0.95}{
\begin{tabular}{c|c}
\hline
\textbf{Action} & \textbf{Description}                                                             \\ \hline
Forward         & Move forward. Press "W" in keyboard.                                                 \\ \hline
Back            & Move backward. Press "S" in keyboard.                                                \\ \hline
Left            & Move left. Press "A" in keyboard.                                                    \\ \hline
Right           & Move right. Press "D" in keyboard.                                                   \\ \hline
Jump            & Jump up. Press "Space" in keyboard.                    \\ \hline
Toggle Inventory            & Open or close the inventory GUI. Press "E" in keyboard.                    \\ \hline
Attack          & Attack items. Click left button in mouse.                                          \\ \hline
Use             & Use items in front of the agent. Click right button in mouse.         \\ \hline
Yaw             & Move agent's view in the horizontal direction. \\ \hline
Pitch           & Move agent's view in the vertical direction. \\ \hline
Hotbar          & Select the item in the hotbar to equip it to the agent's hand. Press 1-9 in keyboard.           \\ \hline
\end{tabular}
}
\label{tab:action_space}
\end{table*}

\begin{table*}[!htb]
\caption{Details of the encapsulation of used steps in our experiments.}
\scalebox{0.95}{
\begin{tabular}{c|p{14cm}}
\hline
\textbf{Step} & \multicolumn{1}{c}{\textbf{Description}}                                                                                                                                                                                                                                                                              \\ \hline
Craft         & Take the crafting table from the inventory. Place it on the ground. Open the crafting table, select the required materials based on the recipe to craft the target item. Take the crafted target item. After that, close the crafting table, destroy the crafting table, and return it to the inventory. \\ \hline
Smelt         & Take the furnace from the inventory. Place it on the ground. Open the furnace, select the required materials based on the recipe to smelt the target item. Take the smelted target item. After that, close the furnace, destroy the furnace, and return it to the inventory.                    \\ \hline
Equip         & Locate the cell of the target item in the inventory or hotbar and equip it to the agent's hand.                                                                                                                                                                                                                                   \\ \hline
Mine          & Directly execute the natural language instructions using the SOTA low-level controller model MineDreamer~\cite{zhou2024minedreamer}.                                                                                                                                                                                                                               \\ \hline
Dig down      & Directly execute the natural language instructions using the SOTA low-level controller model MineDreamer~\cite{zhou2024minedreamer}.                                                                                                                                                                                                                                                                                           \\ \hline
\end{tabular}
}
\label{tab:step_space}
\end{table*}

\subsubsection{Environment Setting}
In dynamic planning experiments, to make sure the comparison between different methods is fair and the agent can successfully execute the generated plans, we need to control the environment to remove the influence of irrelevant factors in the environment. For example, for the task ``Craft 1 \textimage{Icons/diamond_pickaxe.png}~(diamond pickaxe)'', when executing the step ``Mine 3 \textimage{Icons/diamond.png}~(diamond)'', the agent may not be able to mine enough diamonds even if the agent uses the right tool to mine in below the ground because the storage of diamonds below the ground is too scarce. 

To solve this problem, we conduct two types of initialization in the environment:

\noindent
\textbf{Global Environment Initialization.} The rules used to set global setting are listed as follows:
\begin{itemize}
    \item We set the agent to be born in a fixed position in a fixed scene with an empty inventory each time. Therefore, the agent executes the plan starting from a fixed initial state every time. 
    \item We set the agent to have a bright field of view even in the dark. Therefore, the execution of steps will not be affected by the darkness when the agent is below the ground. 
    \item We set the environment to peaceful mode so that no other creatures (such as zombies) can interfere with the agent to execute the plans. 
    \item We set that the environment is always in daylight. So the agent can better execute the plans on above the ground with adequate lighting. 
\end{itemize}

\noindent
\textbf{Ore Distribution Initialization.} To ensure that there is sufficient ore below the ground for the agent to mine, we place some ore below the ground when initializing the environment, so that the agent is more likely to mine the required ore as long as the plans are correct. The command used to place the ore is listed as follows:

\begin{tcolorbox}[colback=prompt_color, colframe=white, boxrule=0pt,breakable]
/fill <from\_x> <from\_y> <from\_z> <to\_x> <to\_y> <to\_z> minecraft:ore
\end{tcolorbox}

Specifically, we implemented an ore distribution algorithm to place the ore. This algorithm generates multiple concentric square layers of ore around the agent's born position in below the ground. In this way, there will be enough ore below the ground for the agent to mine.

\section{Details of Our Agent}
\label{app:our_method}
We list the prompts used in our designed agent as follows.

To conduct the goal decomposition by LLM, we prompt the LLM to decompose each goal into step and sub-goals. For a fair comparison, keeping consistent with other baselines, we only provide the recipe information of the target item required by the task goal, excluding the intermediate sub-goals. Prompt used for goal decomposition in section~\ref{sec:decomposition} is as follows:
\begin{tcolorbox}[colback=prompt_color, colframe=white, boxrule=0pt,breakable]
\textbf{System:}\\
You are a helpful assistant in Minecraft. I will give you a goal to achieve in Minecraft, and you need to decompose the goal into a single step and a list of sub goals to achieve. Output the reasoning thought and the decomposed result. You can follow the history dialogue to make the decomposition.\\

\textbf{User:}\\
========\\
Goal: collect 3 stone.\\
Thought:\\
To collect 3 stone, the last step is to mine 3 stone with wooden\_pickaxe, as mining the stone requires at least wooden\_pickaxe. And the previous sub goals are to obtain 1 wooden\_pickaxe and dig down with wooden\_pickaxe, because stone only appears at the below ground level. Based on these analysis, the decomposed result is as follows:\\
Decomposed Step:\\
Mine 3 stone with wooden\_pickaxe\\
Decomposed Sub Goals:\\
1. Obtain 1 wooden\_pickaxe\\
2. Dig down with wooden\_pickaxe\\
========\\

========\\
Goal: \{new goal\}\\
Thought:\\

\textbf{Assistant:}\\
\end{tcolorbox}

Prompt used to rate for all the steps in the initial plan in section~\ref{sec:consistency} is as follows:
\begin{tcolorbox}[colback=prompt_color, colframe=white, boxrule=0pt,breakable]
\textbf{System:}\\
You are a helpful assistant in Minecraft. I will give you a goal to achieve in Minecraft and the generated initial plan to achieve this goal. You need to rate for all the steps in the initial plan to evaluate the correctness of the steps on a scale of 1 to 10, where 1 indicating the next few steps starting from this step is likely to be wrong and 10 indicating the next few steps starting from this step is completely accurate. \\

\textbf{User:}\\
========\\
Goal: collect 3 stone.\\
Initial Plan:\\
1. Mine 3 log with barehand\\
2. Craft 9 planks\\
3. Craft 2 stick\\
4. Craft 1 crafting\_table\\
5. Craft 1 wooden\_pickaxe\\
6. Mine 3 stone with wooden\_pickaxe\\
Thought:\\
As the stone only exists in below the ground, when executing the step ``Mine 3 stone with wooden\_pickaxe'', the agent should be below the ground. However, before this step the agent is gathering materials and crafting items above the ground. So the latter half of the steps in the initial plan may not be executed successfully.\\
Rating:\\
1. Mine 3 log with barehand - 10\\
2. Craft 9 planks - 8\\
3. Craft 2 stick - 5\\
4. Craft 1 crafting\_table - 3\\
5. Craft 1 wooden\_pickaxe - 3\\
6. Mine 3 stone with wooden\_pickaxe - 5\\
========\\

========\\
Goal: \{goal of new task\}\\
Initial Plan:\\
\{generated initial plan\}\\
Thought:\\

\textbf{Assistant:}\\
\end{tcolorbox}

Prompt used to complement the partial plan between the given start and end anchor steps in section~\ref{sec:consistency} is as follows:
\begin{tcolorbox}[colback=prompt_color, colframe=white, boxrule=0pt,breakable]
\textbf{System:}\\
You are a helpful assistant in Minecraft. I will give you a task goal to achieve in Minecraft and a pair of start and end anchor steps chosen from the corresponding plan to achieve the task goal. You need to complement the partial plan between the start and end anchor steps to help achieve the task goal.\\

\textbf{User:}\\
========\\
Goal: collect 3 stone.\\
Start Anchor Step: Craft 1 crafting\_table\\
End Anchor Step: Mine 3 stone with wooden\_pickaxe\\

Thought:\\
To achieve the task goal ``collect 3 stone'', starting from the step ``Craft 1 crafting\_table'', next I need to craft 1 wooden\_pickaxe to mine stone. But before mining stone, I need to reach below ground first as stone only exists underground. Finally I can mine 3 stone with wooden\_pickaxe. Based on these analysis, the partial plan should be as follows:\\
Partial Plan:\\
Craft 1 crafting\_table\\
Craft 1 wooden\_pickaxe\\
Dig down with wooden\_pickaxe\\
Mine 3 stone with wooden\_pickaxe\\
========\\

========\\
Goal: \{goal of new task\}\\
Start Anchor Step: \{given start anchor step\}\\
End Anchor Step: \{given end anchor step\}\\

Thought:\\

\textbf{Assistant:}\\
\end{tcolorbox}

Prompt used to integrate the initial plan and the complementary partial plan is as follows:
\begin{tcolorbox}[colback=prompt_color, colframe=white, boxrule=0pt,breakable]
\textbf{System:}\\
You are a helpful assistant in Minecraft. I will give you an initial plan to achieve a goal in Minecraft and a complementary partial plan. You need to compare the two plans and correct possible mistakes in the initial plan.\\

\textbf{User:}\\
========\\
Goal: collect 3 stone.\\

Initial Plan:\\
1. Mine 3 log with barehand\\
2. Craft 9 planks\\
3. Craft 2 stick\\
4. Craft 1 crafting\_table\\
5. Craft 1 wooden\_pickaxe\\
6. Mine 3 stone with wooden\_pickaxe\\

Complementary Partial Plan:\\
4. Craft 1 crafting\_table\\
5. Craft 1 wooden\_pickaxe\\
6. Dig down with wooden\_pickaxe\\
7. Mine 3 stone with wooden\_pickaxe\\

Thought:\\
The initial plan ignores one important step ``Dig down with wooden\_pickaxe'' that is presented in the complementary partial plan. As the stone is in below ground, the agent should reach below ground first and then mine the stone.\\

Corrected Plan:\\
1. Mine 3 log with barehand\\
2. Craft 9 planks\\
3. Craft 2 stick\\
4. Craft 1 crafting\_table\\
5. Craft 1 wooden\_pickaxe\\
6. Dig down with wooden\_pickaxe\\
7. Mine 3 stone with wooden\_pickaxe\\
========\\

========\\
Goal: \{goal of new task\}\\

Initial Plan:\\
\{generated initial plan\}\\

Complementary Partial Plan:\\
\{complementary partial plan\}\\

Thought:\\

\textbf{Assistant:}\\
\end{tcolorbox}

\section{Details of Baseline Models}
\label{app:baseline}
In this section, we will provide the details of the baselines used in our experiments. 




\subsection{Baselines in Static Planning}
Baselines in static planning do not interact with the environment. The plans are generated by prompting LLMs with specific modeling methods in the original studies. We adopt ``Meta-Llama-3-8B-Instruct''~\footnote{\url{https://huggingface.co/meta-llama/Meta-Llama-3-8B-Instruct}} and ``OpenAI-gpt-4'' as the backbone model for ``Chain-of-thought'', ``Reverse Chain'', and ``Self-Refine''. 

\noindent
\textbf{Chain-of-Thought~\cite{wei2022chain}.}
\label{appendix:cot}
Chain-of-Thought utilizes forward reasoning based thought chains to generate the plans step by step. Prompt used for Chain-of-Thought is as follows:

\begin{tcolorbox}[colback=prompt_color, colframe=white, boxrule=0pt,breakable]
\textbf{System:}\\
You are a helpful assistant in Minecraft. I will give you a goal to achieve in Minecraft, and you need to decompose the goal into a sequence of steps to help achieve the goal. Output the reasoning thought and the decomposed result. You can follow the history dialogue to make the decomposition.\\

\textbf{User:}\\
========\\
Goal: collect 3 stone.\\
Recipe: stone can be mined below the ground with at least wooden\_pickaxe.\\

Thought: \\
Now let's think step by step. I should first mine 3 log with barehand. Then use 3 log to craft 9 planks. Then use 2 planks to craft 2 stick. Then use 4 planks to craft 1 crafting\_table. Then use 3 planks and 2 stick to craft 1 wooden\_pickaxe on 1 crafting\_table. Then dig down with wooden\_pickaxe. Then mine 3 stone with wooden\_pickaxe. Based on the above reasoning, the plan to achieve the goal is:\\
Plan:
\\1. Mine 3 log with barehand\\2. Craft 9 planks\\3. Craft 2 stick \\4. Craft 1 crafting\_table\\5. Craft 1 wooden\_pickaxe\\6. Dig down with wooden\_pickaxe\\7. Mine 3 stone with wooden\_pickaxe\\
========\\

========\\
Goal: \{goal of new task\}.\\
Recipe: \{recipe of new task\}.\\

Thought:\\

\textbf{Assistant:}\\
\end{tcolorbox}

\noindent
\textbf{Reverse Chain~\cite{zhang2024reverse}.}
In contrast to Chain-of-thought, Reverse Chain utilizes backward reasoning based thought chains to generate the plans. Prompt used in Reverse Chain is as follows:

\begin{tcolorbox}[colback=prompt_color, colframe=white, boxrule=0pt,breakable]
\textbf{System:}\\
You are a helpful assistant in Minecraft. I will give you a goal to achieve in Minecraft, and you need to decompose the goal into a sequence of steps. Output the reasoning thought and the decomposed result. You can follow the history dialogue to make the decomposition. 
\\

\textbf{User:}\\
========\\
Goal: collect 3 stone.\\
Recipe: stone can be mined below the ground with at least wooden\_pickaxe.\\

Thought: \\
Now let's think step by step by backward reasoning. The final goal is to collect 3 stone. To collect 3 stone, I need to mine 3 stone with wooden\_pickaxe. stone usually exists underground, so I need to dig down with wooden\_pickaxe. To obtain 1 wooden\_pickaxe, I need to craft 1 wooden\_pickaxe using 3 planks and 2 stick on 1 crafting\_table. To obtain 1 crafting\_table, I need to craft 1 crafting\_table using 4 planks. To obtain 2 stick, I need craft 2 stick using 2 planks on 1 crafting\_table. So I totally need 9 planks. To obtain 9 planks, I need to craft 9 planks using 3 log, so I need to mine 3 log first. Mining 3 log can be done with barehand, so I start from here. Based on the above reasoning, the plan to complete this task is:\\
Plan:
\\1. Mine 3 log with barehand\\2. Craft 9 planks\\3. Craft 2 stick \\4. Craft 1 crafting\_table\\5. Craft 1 wooden\_pickaxe\\6. Dig down with wooden\_pickaxe\\7. Mine 3 stone with wooden\_pickaxe\\
========\\

========\\
Goal: \{goal of new task\}.\\
Recipe: \{recipe of new task\}.\\

Thought:\\

\textbf{Assistant:}\\
\end{tcolorbox}

\noindent
\textbf{Self-Refine~\cite{madaan2024self}.}
Self-Refine first generates an initial plan using the same prompt as Chain-of-Thought. The generated initial plan is then sent to the same LLM to obtain feedback. Finally, the feedback and the generated initial plan are sent to the same LLM to produce a refined plan. This process is repeated until the LLM's feedback indicates that the generated plan is totally correct. 

Prompt used to generate the feedback for the already generated plan is as follows:

\begin{tcolorbox}[colback=prompt_color, colframe=white, boxrule=0pt,breakable]
\textbf{System:}\\
You are a helpful assistant in Minecraft. I will give you a goal and the corresponding plan to achieve it in Minecraft, and you need to check the correctness of the plan and output the feedback. You need to output "Correct" or  "Incorrect" first. If outputting "Incorrect", then output the detailed analysis and list the mistakes.\\

\textbf{User:}\\
========\\
Goal: collect 3 stone.\\
Recipe: stone can be mined below the ground with at least wooden\_pickaxe.\\

Plan:\\1. Mine 3 log with barehand\\2. Craft 9 planks\\3. Craft 2 stick\\4. Craft 1 wooden\_pickaxe\\5. Dig down with wooden\_pickaxe\\6.Mine 3 log with wooden\_pickaxe\\7. Mine 3 stone with wooden\_pickaxe\\

What's wrong with the plan? Please give me the feedback and list the mistakes if the plan is wrong.\\

Incorrect. Step 4 "Crafting 1 wooden\_pickaxe" must be done on a crafting\_table. But the plan does not include "Craft 1 crafting\_table" before Step 4, so the first mistake is "ignoring obtaining one crafting\_table before using it to craft other tools". Step 6 "Mine 3 log with wooden\_pickaxe" is executed after Step 5 "Dig down with wooden\_pickaxe". But log only exists on above the ground. So the second mistake is "collecting materials on above the ground when the agent is already in below the ground".\\

Mistake:\\1. Ignoring obtaining one crafting\_table before using it to craft other tools.\\2. Collecting materials on above the ground when the agent is already in below the ground.\\
========\\

========\\
Goal: \{goal of new task\}\\
Recipe: \{recipe of new task\}\\

Plan:\\
\{generated initial plan of new task\}\\

What's wrong with the plan? Please give me the feedback and list the mistakes if the plan is wrong.\\

\textbf{Assistant:}\\
\end{tcolorbox}

Prompt used to refine the initial plan based on the feedback given by LLM:

\begin{tcolorbox}[colback=prompt_color, colframe=white, boxrule=0pt,breakable]
\textbf{System:}\\
You are a helpful assistant in Minecraft. I will give you a goal to achieve in Minecraft, the initial plan of this goal and the feedback about the initial plan. You need to correct the initial plan according to the feedback and output the refined plan.\\

\textbf{User:}\\
========\\
Goal: collect 3 stone.\\
Recipe: stone can be mined below the ground with at least wooden\_pickaxe.\\

Initial Plan:\\1. Mine 3 log with barehand\\2. Craft 9 planks\\3. Craft 2 stick\\4. Craft 1 wooden\_pickaxe\\5. Dig down with wooden\_pickaxe\\6. Mine 3 log with wooden\_pickaxe\\7. Mine 3 stone with wooden\_pickaxe\\

Feedback:\\
Incorrect. Step 4 "Crafting 1 wooden\_pickaxe" must be done on a crafting\_table. But the plan does not include "Craft 1 crafting\_table" before Step 4, so the first mistake is "ignoring obtaining one crafting\_table before using it to craft other tools". Step 6 "Mine 3 log with wooden\_pickaxe" is executed after Step 5 "Dig down with wooden\_pickaxe". But log only exists on above the ground. So the second mistake is "collecting materials on above the ground when the agent is already in below the ground".\\

Mistake:\\1. Ignoring obtaining one crafting\_table before using it to craft other tools.\\2. Collecting materials on above the ground when the agent is already in below the ground.\\

Refined Plan:\\
1. Mine 3 log with barehand\\2. Craft 9 planks\\3. Craft 2 stick \\4. Craft 1 crafting\_table\\5. Craft 1 wooden\_pickaxe\\6. Dig down with wooden\_pickaxe\\7. Mine 3 stone with wooden\_pickaxe\\
========\\

========\\
Goal: \{goal of new task\}\\
Recipe: \{recipe of new task\}\\

Initial Plan:\\
\{initial plan of new task\}\\

Feedback:\\
\{feedback of the initial plan\}\\

Refined Plan:\\

\textbf{Assistant:}\\
\end{tcolorbox}

\noindent
\textbf{Tree-of-thought~\cite{yao2024tree}.}
Tree-of-thought generates the plan for each task by exploring multiple possible paths simultaneously. Starting from the initial state, Tree-of-thought samples three possible steps each time that can be executed next. In our experiments, we adopt ``OpenAI-gpt-4'' as the backbone model. We set the temperature of LLM to $0.7$ and sample for three times to get three possible next step each time. During planning, Tree-of-thought maintain multiple generated plans and select the best one through LLM rating. The generation of each plan ends when inferring the step to create the required item of the task goal. 

The prompt used to sample possible steps is shown as follows:
\begin{tcolorbox}[colback=prompt_color, colframe=white, boxrule=0pt,breakable]
\textbf{System:}\\
You are a helpful assistant in Minecraft. I will give you a goal to achieve in Minecraft, the recipe for the required item of the goal and the already generated partial plan. You need to generate the next step for current plan. 
\\

\textbf{User:}\\
========\\
Goal: collect 3 stone.\\
Recipe: stone can be mined below the ground with at least wooden\_pickaxe.\\
Already Generated Plan:\\1. Mine 3 log with barehand\\2. Craft 9 planks\\3. Craft 2 stick\\
Next Step:\\
4. Craft 1 crafting\_table\\
========\\

========\\
Goal: \{goal of new task\}\\
Recipe: \{recipe of new task\}\\
Already Generated Plan:\\\{generated partial plan\}\\
Next Step:\\

\textbf{Assistant:}\\
\end{tcolorbox}

After generating multiple candidate plans, the best plan is selected by taking the average result of three times of rating by LLMs. The prompt used to rate generated plan is shown as follows:
\begin{tcolorbox}[colback=prompt_color, colframe=white, boxrule=0pt,breakable]
\textbf{System:}\\
Evaluate if the given plan can help to reach the goal (sure/likely/impossible)
\\

\textbf{User:}\\
========\\
Goal: obtain 1 wooden\_pickaxe.\\
Plan:\\1. Mine 3 log with barehand\\2. Craft 9 planks\\3. Craft 2 stick\\4. Craft 1 crafting\_table\\5. craft 1 wooden\_pickaxe\\
Evaluation:\\
sure\\
========\\

========\\
Goal: obtain 1 wooden\_pickaxe.\\
Plan:\\1. Mine 3 log with barehand\\2. Craft 9 planks\\3. Craft 2 stick\\4. craft 1 wooden\_pickaxe\\
Evaluation:\\
impossible\\
========\\

========\\
Goal: \{goal of new task\}\\
Plan: \{generated plan of new task\}\\
Evaluation:\\

\textbf{Assistant:}\\
\end{tcolorbox}

\noindent
\textbf{DEPS~\cite{wang2023describe}.}
DEPS first generates the initial plan and then re-generates the plan through integrating the description and the explanation of the plan given by large language models. We adopt ``OpenAI-gpt-4'' as the backbone model. 

DEPS first generates the initial plan by in-context learning in a code style:

\begin{tcolorbox}[colback=prompt_color, colframe=white, boxrule=0pt,breakable]
\textbf{System:}\\
You are a helper agent in Minecraft. You need to generate the sequences of steps for a certain task in Minecraft.\\

\textbf{User:}\\
========\\
Goal: collect 3 stone.\\
Recipe: stone can be mined below the ground with at least wooden\_pickaxe.\\
The code for collecting 3 stone is as follows:\\
def collect\_3\_stone(\hspace{3pt}):\\
\textcolor{prompt_color}{1} \hspace{5pt} mine(\{’log’:3\}, null);\\
\textcolor{prompt_color}{1} \hspace{5pt} craft(\{’planks’:9\}, null);\\
\textcolor{prompt_color}{1} \hspace{5pt} craft(\{’stick’:2\}, null);\\
\textcolor{prompt_color}{1} \hspace{5pt} craft(\{’crafting\_table’:1\}, null);\\
\textcolor{prompt_color}{1} \hspace{5pt} craft(\{’wooden\_pickaxe’:1\}, null);\\
\textcolor{prompt_color}{1} \hspace{5pt} dig\_down(wooden\_pickaxe);\\
\textcolor{prompt_color}{1} \hspace{5pt} mine(\{’stone’:3\}, wooden\_pickaxe);\\
\textcolor{prompt_color}{1} \hspace{5pt} return `3 stone';\\
========\\

========\\
Goal: \{goal of new task\}\\
Recipe: \{recipe of new task\}\\

\textbf{Assistant:}\\
\end{tcolorbox}

Then, DEPS fixes the possible mistakes in the initial plan based on the description and the explanation of the initial plan:

\begin{tcolorbox}[colback=prompt_color, colframe=white, boxrule=0pt,breakable]
\textbf{System:}\\
You are a helper agent in Minecraft. You need to generate the sequences of steps for a certain task in Minecraft.\\

\textbf{User:}\\
========\\
Goal: collect 3 stone.\\
Recipe: stone can be mined below the ground with at least wooden\_pickaxe.\\
The code for collecting 3 stone is as follows:\\
def collect\_3\_stone(\hspace{3pt}):\\
\textcolor{prompt_color}{1} \hspace{5pt} mine(\{’log’:3\}, null);\\
\textcolor{prompt_color}{1} \hspace{5pt} craft(\{’planks’:9\}, null);\\
\textcolor{prompt_color}{1} \hspace{5pt} craft(\{’stick’:2\}, null);\\
\textcolor{prompt_color}{1} \hspace{5pt} craft(\{’wooden\_pickaxe’:1\}, null);\\
\textcolor{prompt_color}{1} \hspace{5pt} dig\_down(wooden\_pickaxe);\\
\textcolor{prompt_color}{1} \hspace{5pt} mine(\{’stone’:3\}, wooden\_pickaxe);\\
\textcolor{prompt_color}{1} \hspace{5pt} return `3 stone';\\
Descriptor: The step ``craft(\{’wooden\_pickaxe’:1\}, null)'' will not be executed successfully.\\
Explainer: Because I need a crafting\_table to craft the wooden\_pickaxe.\\
RePlanner: The code for collecting 3 stone is as follows:\\
def collect\_3\_stone(\hspace{3pt}):\\
\textcolor{prompt_color}{1} \hspace{5pt} mine(\{’log’:3\}, null);\\
\textcolor{prompt_color}{1} \hspace{5pt} craft(\{’planks’:9\}, null);\\
\textcolor{prompt_color}{1} \hspace{5pt} craft(\{’stick’:2\}, null);\\
\textcolor{prompt_color}{1} \hspace{5pt} craft(\{’crafting\_table’:1\}, null);\\
\textcolor{prompt_color}{1} \hspace{5pt} craft(\{’wooden\_pickaxe’:1\}, null);\\
\textcolor{prompt_color}{1} \hspace{5pt} dig\_down(wooden\_pickaxe);\\
\textcolor{prompt_color}{1} \hspace{5pt} mine(\{’stone’:3\}, wooden\_pickaxe);\\
\textcolor{prompt_color}{1} \hspace{5pt} return `3 stone';\\
========\\

========\\
Goal: \{goal of new task\}\\
Recipe: \{recipe of new task\}\\
{generated code for new task}

\textbf{Assistant:}\\
\end{tcolorbox}

\noindent
\textbf{Plan-and-Solve~\cite{wang2023plan}.}
Plan-and-Solve enables the large language model to generate the intermediate planning results before outputting the final plan. Plan-and-Solve can calculate the execution results of generated plans by in-context learning and thus enhances the accuracy of generated plans. We adopt ``OpenAI-gpt-4'' as the backbone model. The prompt used to generate the plans is shown as follows:

\begin{tcolorbox}[colback=prompt_color, colframe=white, boxrule=0pt,breakable]
\textbf{System:}\\
You are a helpful assistant in Minecraft. I will give you a goal to achieve in Minecraft, the recipe for the required item of the goal. You need to generate the plan to achieve the task goal.\\

\textbf{User:}\\
========\\
Goal: collect 3 stone.\\
Recipe: stone can be mined below the ground with at least wooden\_pickaxe.\\

Let's first understand the task and devise a plan to achieve the task goal.\\
Then, let's carry out the plan and achieve the task goal step by step.\\

Plan:\\
1. Mine 3 log with barehand\\
2. Craft 9 planks\\
3. Craft 2 stick\\
4. Craft 1 crafting\_table\\
5. Craft 1 wooden\_pickaxe\\
6. Dig down with wooden\_pickaxe\\
7. Mine 3 stone with wooden\_pickaxe\\

Task goal achieving:\\
1. I obtain 3 log by mining with barehand\\
2. I obtain 9 planks using 3 log\\
3. I obtain 2 stick using 2 planks, I have 7 planks left\\
4. I obtain 1 crafting\_table using 4 planks, I have 3 planks left\\
5. I obtain 1 wooden\_pickaxe using 2 stick and 3 planks on 1 crafting\_table using, I have 0 planks left and 0 stick left\\
6. I reach below the ground by digging down with wooden\_pickaxe\\
7. I obtain 3 stone by mining with wooden\_pickaxe\\

In conclusion, I have achieved the task goal by obtaining 3 stone.\\
========\\

========\\
Goal: \{goal of new task\}\\
Recipe: \{recipe of new task\}\\

Let's first understand the task and devise a plan to achieve the task goal.\\
Then, let's carry out the plan and achieve the task goal step by step.\\

Plan:\\

\textbf{Assistant:}\\
\end{tcolorbox}

\noindent
\textbf{Openai-o1~\cite{o1_2024}.}
Openai-o1 uses the same prompt as Chain-of-thought. However, Openai-o1 will conduct more exploration on different paths and more refinements before outputting the final plans. 


\subsection{Baselines in Dynamic Planning}
In dynamic planning experiments, the agent needs to interact with the environment and obtain the feedback to help conduct the planning. Therefore, the baseline models need to inform LLMs the feedback from the environment through prompts and utilize LLMs' reasoning ability to help generate proper plans. 

\noindent
\textbf{React~\cite{yao2023react}.}
\label{section:ReAct}
React first conduct reasoning based on the task goal and the observation from the environment to infer the next few steps to be executed. Then it executes the inferred steps in the environment to obtain new observation, and repeats the reasoning and execution process until achieving the task goal. We adopt ``OpenAI-gpt-4'' as the backbone model. The used prompt is shown as follows:
\begin{tcolorbox}[colback=prompt_color, colframe=white, boxrule=0pt,breakable]
\textbf{System:}\\
You are a helpful assistant in Minecraft. I will give you a goal to achieve in Minecraft, and you need to reason what steps to be executed next based on the task goal and the observation from the environment. Output the reasoning thought and the next few steps. After executing these steps in the environment, you will receive the observation from the environment. Then you need to reflect the execution results and continue to output the next few steps to be executed.\\

========\\
\textbf{User:}\\
Goal: collect 3 stone.\\
Recipe: stone can be mined below the ground with at least wooden\_pickaxe.\\

\textbf{Assistant:}\\
Reasoning:\\
There is nothing in my inventory. I need to collect stone, so I first need to craft a wooden\_pickaxe, arrive at below the ground, and mine the stone. Therefore, I need to craft the wooden\_pickaxe. \\
Steps:\\
Mine 2 log with barehand\\
Craft 8 planks\\
Craft 2 stick\\
Craft 1 crafting\_table\\
Craft 1 wooden\_pickaxe\\

\textbf{User:}\\
Observation:\\
Mine 2 log with barehand \textcolor{blue}{completed}, there are 2 log in my inventory. Craft 8 planks \textcolor{blue}{completed}, there are 0 log, 8 planks in my inventory. Craft 2 stick \textcolor{blue}{completed}, there are 6 planks, 2 stick in my inventory. Craft 1 crafting\_table \textcolor{blue}{completed}, there are 2 planks, 2 stick, 1 crafting\_table in my inventory. Craft 1 wooden\_pickaxe \textcolor{red}{failed}.\\

\textbf{Assistant:}\\
Reasoning:\\
I \textcolor{red}{failed} crafting 1 wooden\_pickaxe since there is no enough planks in my inventory. So I need to mine 1 more log to craft more planks. Then execute the following steps:\\
Steps:\\
Mine 1 log with barehand\\
Craft 1 planks\\
Craft 1 wooden\_pickaxe\\

\textbf{User:}\\
Observation:\\
Mine 1 log with barehand \textcolor{blue}{completed}, there are 1 log, 2 planks, 1 crafting\_table, 2 stick in my inventory. Craft 1 planks \textcolor{blue}{completed}, there are 0 log, 3 planks, 1 crafting\_table, 2 stick in my inventory. Craft 1 wooden\_pickaxe \textcolor{blue}{completed}, there are 1 crafting\_table, 1 wooden\_pickaxe in my inventory.\\

\textbf{Assistant:}\\
Reasoning:\\
I have 1 wooden\_pickaxe in my inventory now, I can dig down to arrive at below the ground to mine stone. Therefore I need to dig down with wooden\_pickaxe and mine stone]\\
Steps:\\
Dig down with wooden\_pickaxe\\
Mine 3 stone with wooden\_pickaxe\\

\textbf{User:}\\
Observation:\\
Dig down with wooden\_pickaxe \textcolor{blue}{completed}, there are 1 crafting\_table, 1 wooden\_pickaxe in my inventory. I am already below the ground. Mine 3 stone with wooden\_pickaxe \textcolor{blue}{completed}, there are 1 crafting\_table, 1 wooden\_pickaxe, 3 stone in my inventory. \textcolor{achieve}{Goal achieved}.\\
========\\

========\\
\textbf{User:}\\
Goal: \{goal of new task\}\\
Recipe: \{recipe of new task\}\\

\textbf{Assistant:}\\
\end{tcolorbox}

\noindent
\textbf{Jarvis-1~\cite{wang2023jarvis}.}
\label{section:Jarvis-1}
Jarvis-1 can adjust the plans based on the execution results from the environment. Through deep thinking, it can find the reason for the failure of previous plan and generate the corrected plan based on the thinking. We adopt ``OpenAI-gpt-4'' as the backbone model. The prompt used by Jarvis-1 is shown as follows: 
\begin{tcolorbox}[colback=prompt_color, colframe=white, boxrule=0pt,breakable]
\textbf{System:}\\
You are a helpful assistant in Minecraft. I will give you a goal to achieve in Minecraft and the initial plan to achieve the goal. After executing the plan in the environment, you will receive the observation about the execution results of the plan from the environment. You need to reflect the reason for the failure if the plan cannot be executed successfully and output corrected plan based on your reflection. Repeating the above process until the goal being achieved.\\ 

========\\
\textbf{User:}\\
Goal: collect 3 stone.\\
Recipe: stone can be mined below the ground with at least wooden\_pickaxe.\\

Initial Plan:\\1. Mine 2 log with barehand\\2. Craft 8 planks\\3. Craft 2 stick\\4. Craft 1 crafting\_table\\5. Craft 1 wooden\_pickaxe\\6. Mine 3 stone with wooden\_pickaxe\\

Observation:\\
Mine 2 log with barehand \textcolor{blue}{completed}, there are 2 log in my inventory. Craft 8 planks \textcolor{blue}{completed}, there are 0 log, 8 planks in my inventory. Craft 2 stick \textcolor{blue}{completed}, there are 6 planks, 2 stick in my inventory.  Craft 1 crafting\_table \textcolor{blue}{completed}, there are 2 planks, 2 stick, 1 crafting\_table in my inventory. Craft 1 wooden\_pickaxe \textcolor{red}{failed}.\\

\textbf{Assistant:}\\
Thinking:\\
I \textcolor{red}{failed} crafting 1 wooden\_pickaxe since there is no enough planks in my inventory. So I need to mine 1 more log to craft more planks. 

Corrected Plan:\\
1. Mine 3 log with barehand\\
2. Craft 9 planks\\
3. Craft 2 stick\\
4. Craft 1 crafting\_table\\
5. Craft 1 wooden\_pickaxe\\
6. Mine 3 stone with wooden\_pickaxe\\

\textbf{User:}\\
Observation:\\
Mine 3 log with barehand \textcolor{blue}{completed}, there are 3 log in my inventory. Craft 9 planks \textcolor{blue}{completed}, there are 0 log, 9 planks in my inventory. Craft 2 stick \textcolor{blue}{completed}, there are 7 planks, 2 stick in my inventory. Craft 1 crafting\_table \textcolor{blue}{completed}, there are 3 planks, 2 stick, 1 crafting\_table in my inventory. Craft 1 wooden\_pickaxe \textcolor{blue}{completed}, there are 1 crafting\_table, 1 wooden\_pickaxe in my inventory. \textcolor{achieve}{Goal achieved}.\\
========\\

========\\
\textbf{User:}\\
Goal: \{goal of new task\}\\
Recipe: \{recipe of new task\}\\

Initial Plan:\\
\{initial plan of new task\}\\

Observation:\\
\{observation from the environment\}\\

\textbf{Assistant:}\\
\end{tcolorbox}

\begin{figure}[!th]
    \centering
    \includegraphics[width=.45\textwidth]{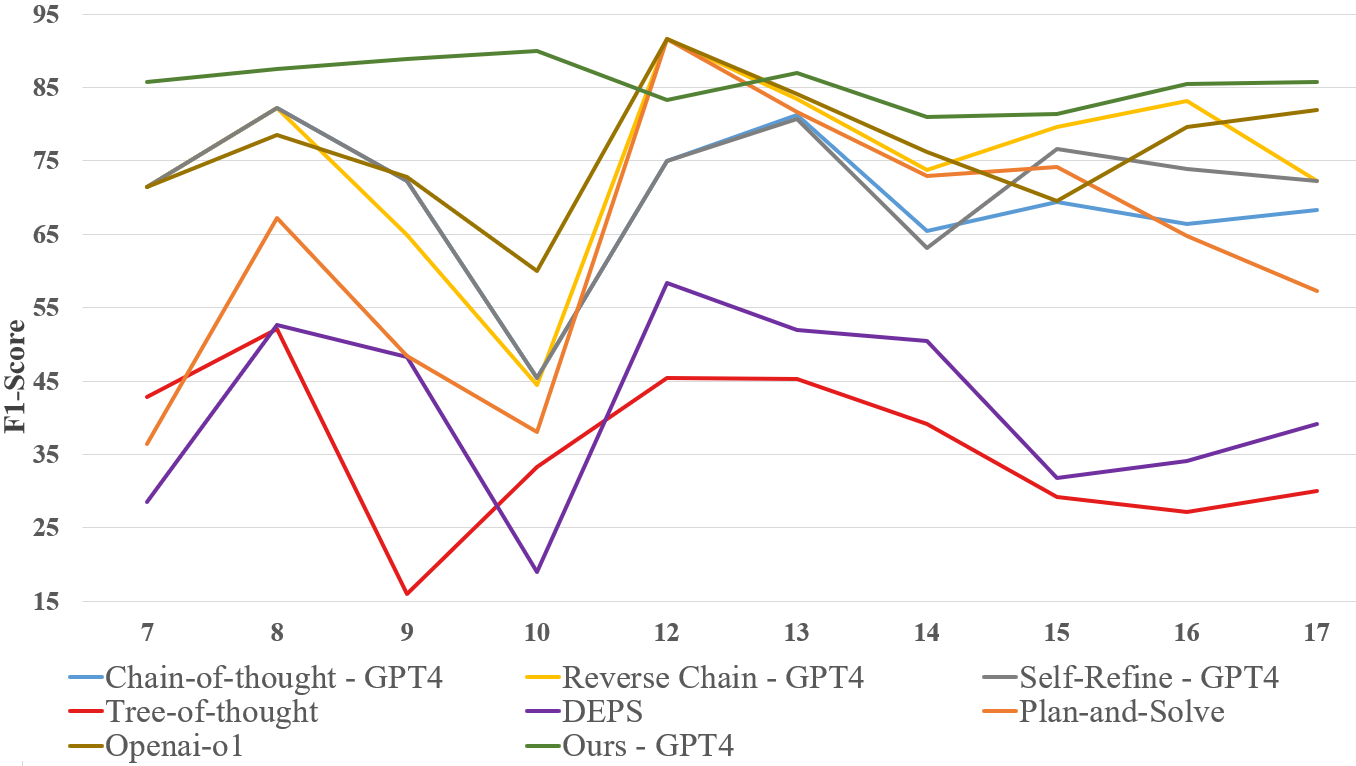}
    \caption{The performance of different models on plans of different lengths. Best viewed on screen.}
    \label{fig:length}
\end{figure}

\begin{figure}[!th]
    \centering
    \includegraphics[width=.45\textwidth]{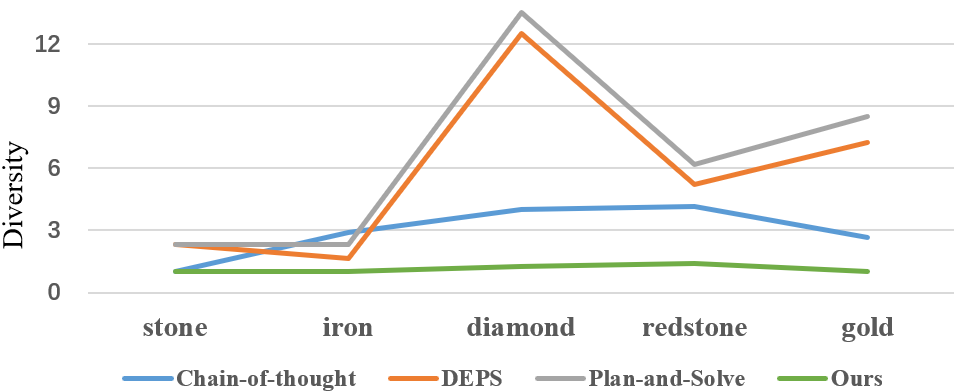}
    \caption{Sample diversity of different models on tasks in different groups. Best viewed on screen.}
    \label{fig:diversity}
\end{figure}

\section{Supplementary Experiments}

\subsection{Advantage Analysis}
To demonstrate the advantages of our proposed agent over other baselines, we evaluate the sensitivity of different models to plan length and the stability of inferring intermediate steps during planning. 
We first evaluate the planning performance of different models on plans of different lengths. As shown in Figure~\ref{fig:length}, the performance of baseline models varies significantly with the increase of the plan length, exhibiting extreme instability, especially for plans from the length of 7 to the length of 12. In contrast, our agent shows no significant variation in performance when the length of the plans increases. This is because our agent can narrow down the situations that the agent needs to consider for goal decomposition through the recursive goal decomposition module. Therefore, the difficulty of goal decomposition will not increase when the plans become longer, enabling our agent to conduct robust planning for various tasks. 

Then we evaluate the stability of immediate step inference for different models. For tasks in each group, we provide a partial plan in the ground truth and require the model to generate the next three steps. For each task, we sample 50 generations for each model and calculate how many different three steps there are in these 50 generations, denoted as diversity value. As shown in Figure~\ref{fig:diversity}, compared to other baselines, our agent consistently maintains low diversity for the planning of different tasks. This demonstrates the stability of our agent in goal decomposition and plan generation. 

\begin{figure}[!ht]
    \centering
    \includegraphics[width=.45\textwidth]{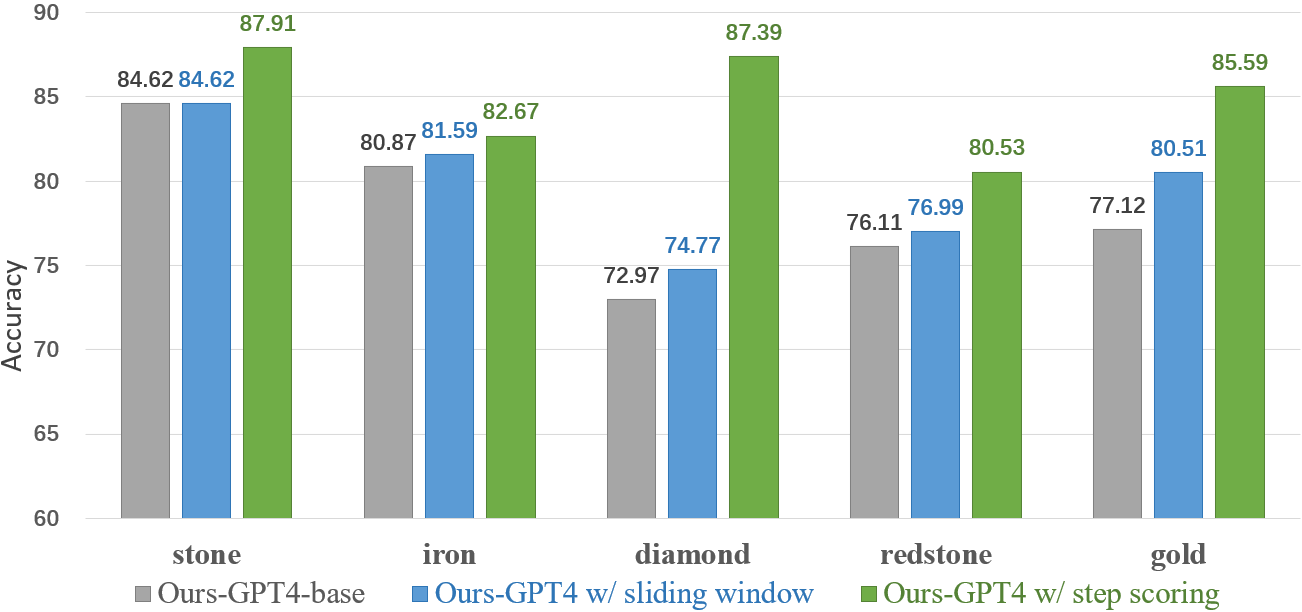}
    \caption{Results of ablation study for state consistency maintaining, GPT4 as the backbone model, under the Accuracy metric.}
    \label{fig:acc_ablation_gpt4}
\end{figure}

\begin{figure}[!ht]
    \centering
    \includegraphics[width=.45\textwidth]{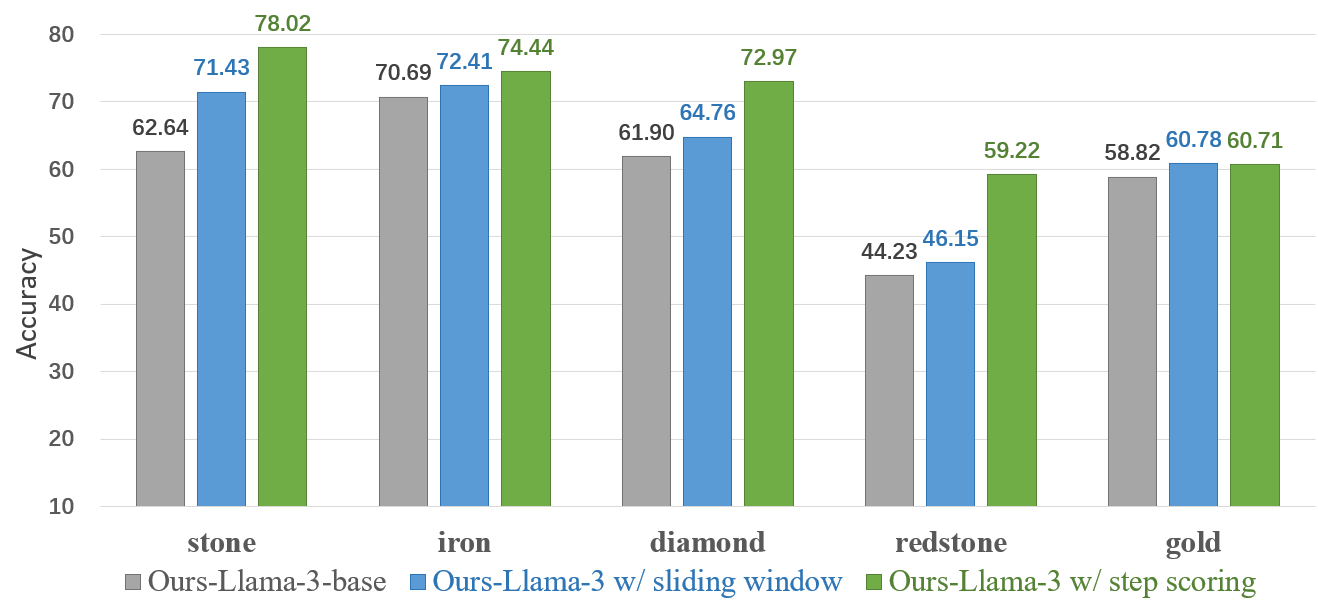}
    \caption{Results of ablation study for state consistency maintaining, Llama-3-8B as the backbone model, under the Accuracy metric.}
    \label{fig:acc_ablation_llama3}
\end{figure}

\begin{figure}[!ht]
    \centering
    \includegraphics[width=.45\textwidth]{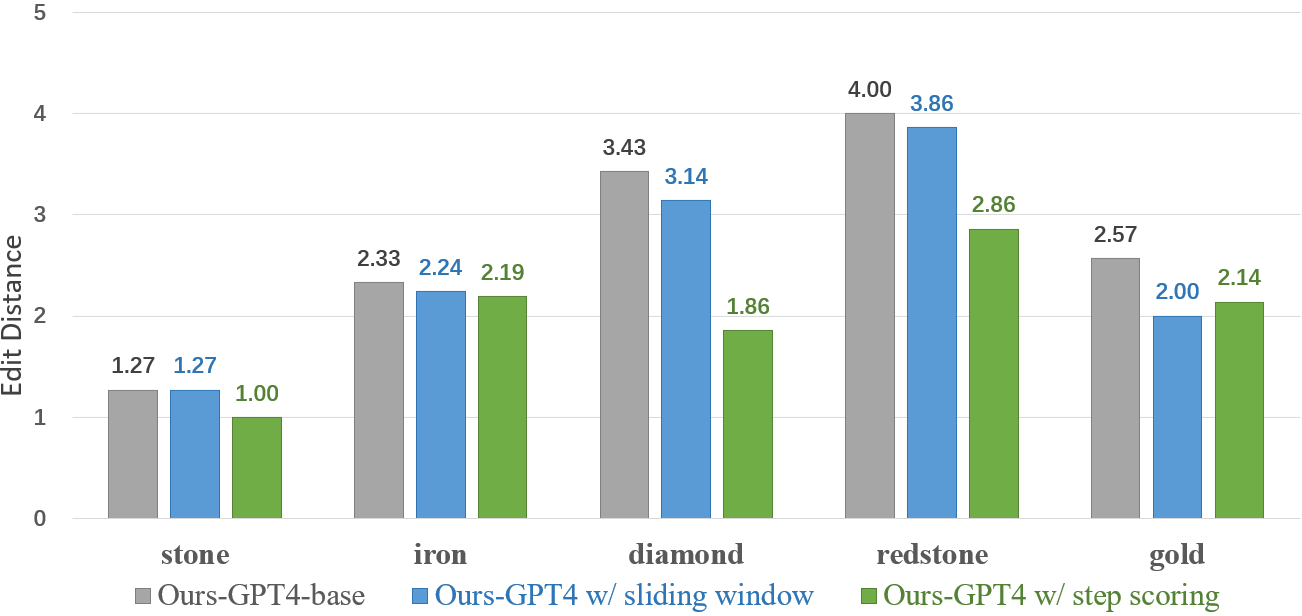}
    \caption{Results of ablation study for state consistency maintaining, GPT4 as the backbone model, under the Edit Distance metric.}
    \label{fig:ed_ablation_gpt4}
\end{figure}

\begin{figure}[!ht]
    \centering
    \includegraphics[width=.45\textwidth]{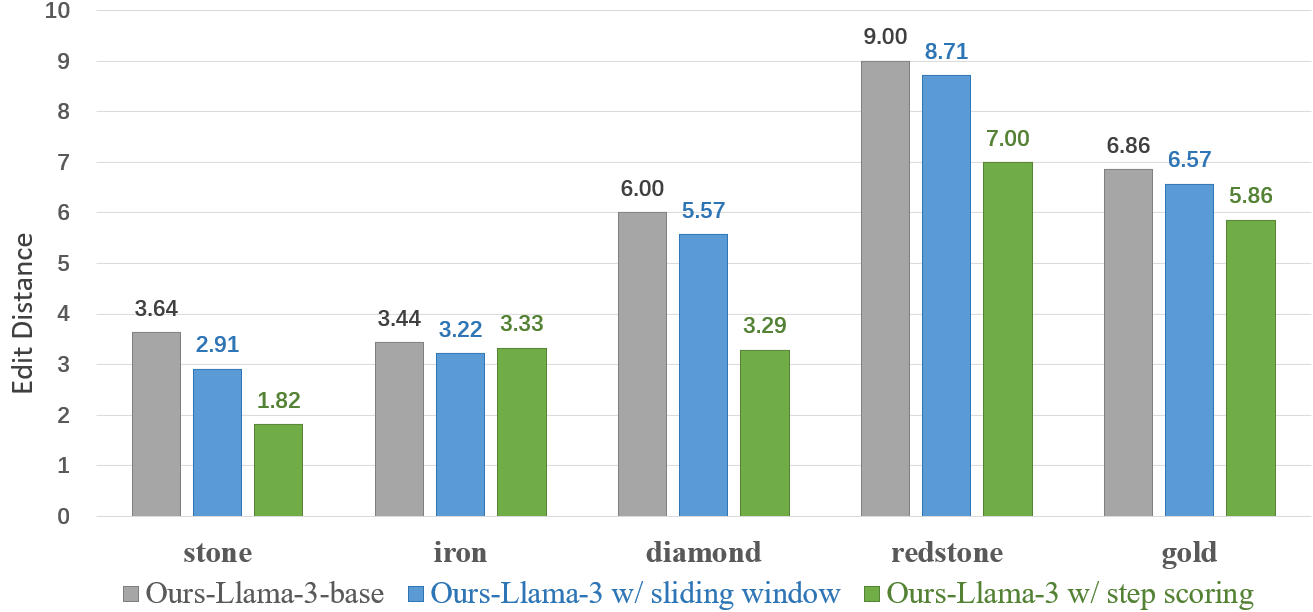}
    \caption{Results of ablation study for state consistency maintaining, Llama-3-8B as the backbone model, under the Edit Distance metric.}
    \label{fig:ed_llama3}
\end{figure}

\subsection{Supplementary Ablation Study}
We show part of the experimental results here that are not presented in the main part due to space limitations. 
For the ablation study in section~\ref{sec:ablation}, in addition to F1-Score, we also present the results under Accuracy and Edit Distance metrics. As shown in Figure~\ref{fig:acc_ablation_gpt4} and Figure~\ref{fig:acc_ablation_llama3}, under the Accuracy metric, our proposed state consistency maintaining module can significantly improve the quality of generated plans. Moreover, As shown in Figure~\ref{fig:ed_ablation_gpt4} and Figure~\ref{fig:ed_llama3}, under the Edit Distance metric, we can still draw the same conclusion because the proposed module can reduce the difference between generated plans and the ground truth. In addition, for the two methods to select anchor steps, the step scoring method can bring greater performance improvement, as the sliding window method chooses too many anchor step pairs and brings unnecessary noise to the plans.

\end{document}